\documentclass[11pt]{article}

\usepackage[letterpaper,margin=1in]{geometry}
\usepackage{times}
\usepackage{microtype}
\usepackage{graphicx}
\usepackage{booktabs} %
\usepackage{array}
\usepackage{amsmath}
\allowdisplaybreaks
\usepackage{amssymb}
\usepackage{bm}
\usepackage{multirow}
\usepackage{placeins}
\usepackage{calc}
\usepackage{pgfplots}
\pgfplotsset{compat=1.9}
\usepackage{algorithm}
\usepackage{algorithmic}

\usepackage{natbib}
\setcitestyle{authoryear,round,citesep={;},aysep={,},yysep={;}}
\renewcommand{\cite}[1]{\citep{#1}}

\usepackage[hypertexnames=false]{hyperref}

\usepackage{xurl}

\newcommand{\schreps}{0.0579847}

\usepackage{xcolor}
\definecolor{matrixbg}{RGB}{252,222,222}
\definecolor{vectorbg}{RGB}{220,242,226}
\newcommand{\unitbox}[2]{{\setlength{\fboxsep}{1.6pt}\colorbox{#1}{#2}}}
\newcommand{\matrixline}[1]{\unitbox{matrixbg}{#1}}
\newcommand{\vectorline}[1]{\unitbox{vectorbg}{#1}}

\renewcommand{\algorithmiccomment}[1]{\hfill$\triangleright$~#1}

\newlength{\naturalwd}
\let\scaleboxdown\resizebox
\renewcommand{\resizebox}[3]{%
  \settowidth{\naturalwd}{#3}%
  \ifdim\naturalwd>#1\relax\scaleboxdown{#1}{#2}{#3}\else\mbox{#3}\fi}

\graphicspath{ {./images/} }

\newlength{\gridlab}
\newlength{\gridw}
\newcommand{\gridhead}[1]{\makebox[\gridw]{\scriptsize #1}}
\newcommand{\gridrow}[1]{\makebox[\gridlab]{\rotatebox{90}{\makebox[0.5625\gridw]{\scriptsize #1}}}}
\newcommand{\gridimg}[1]{\includegraphics[width=\gridw]{#1}}

\title{Fused-Softmax for efficient quantized attention}

\author{%
  Tianxing Wu\thanks{Equal contribution.} \quad
  Juuso Korhonen\footnotemark[1] \quad
  Jesus Carabano Bravo\footnotemark[1] \\[2pt]
  Advanced Micro Devices, Inc., Austin, TX, USA \\[2pt]
  \texttt{\{Tianxing.Wu, Juuso.Korhonen, Jesus.CarabanoBravo\}@amd.com}%
}
\date{}

\hypersetup{pdfkeywords={FlashAttention, SageAttention, Softmax, Quantization, Fusion, Diffusion}}

\begin{document}

\maketitle

\begin{abstract}
Low-precision matrix throughput on modern GPUs is growing far faster than vector-arithmetic
throughput. Attention therefore cannot realize the available GEMM performance because softmax has
become the pipeline bottleneck. We introduce Fused-Softmax Attention, a training-free algorithm that reformulates and
approximates softmax to fold its dominant operations into the attention GEMMs, moving work from
vector units to matrix cores. Implemented for the AMD CDNA\texttrademark{}~4 architecture, our
MXFP8, MXFP6 and MXFP4 kernels reach approximately 3.0, 4.0 and 4.6~PFLOP/s on an AMD
Instinct\texttrademark{} MI355X GPU, yielding speedups of $2.1\times$, $2.8\times$ and $3.2\times$
over the BF16 attention. Importantly, the additional error introduced by Fused-Softmax
remains below the error floor set by quantizing the GEMM operands. On Wan2.2-T2V-A14B, a hybrid
MXFP6/4 schedule delivers up to $2.05\times$ end-to-end speedup while preserving the visual quality
of the outputs. The kernels ship in AMD's open-source AITER library, and they powered AMD's MLPerf\textsuperscript{\textregistered} Inference v6.1 Wan2.2 submission, which ran
$1.7\times$ faster than AMD's previous-round result on the same hardware. Our results show that
redesigning softmax to utilize the matrix pipeline enables attention to capitalize on continued
growth in low-precision matrix throughput.
\end{abstract}

\section{Introduction}

Transformer attention~\cite{vaswani2017attention} is a cornerstone algorithm of modern AI models, but its quadratic complexity
in terms of sequence length makes it a computational bottleneck in long-sequence settings. This is particularly relevant in video generation
models~\cite{wan2025wan, kong2024hunyuanvideo, minimaxai2026h3}, where higher resolution and longer videos increase the effective sequence length. 

The FlashAttention family~\cite{dao2022flashattention,dao2023flashattention2,shah2024flashattention3}
has become the standard approach to map the attention algorithm efficiently to modern GPU hardware. In its latest version, FlashAttention-4~\cite{zadouri2026flashattention4},
software pipelining has evolved to overlap the two kinds of work in attention, the matrix multiplies and the vector work of the softmax,
exploiting independent execution units in modern datacenter GPUs. However, the throughput of these execution units
has scaled asymmetrically across recent datacenter GPU generations, leaving vector throughput far behind
and making the softmax the bottleneck.

Related work addresses this bottleneck by reducing the cost and number of the softmax-related vector operations.
FlashAttention-4 uses polynomial approximation for
part of the exponentials and skips most output rescales. Its follow-ups~\cite{zhang2026attnqat,deng2026mpfa}
quantize the matrix operands to FP8 and below but leave the softmax unchanged. Two concurrent works, VC-Attention~\cite{li2026vcattention} and Direct-P from FP4 FlashAttention-4~\cite{hu2026fp4flashattention4}, 
replace the costly exponential with an affine transformation into the quantized probabilities.

In contrast to these works, our proposed method moves most of the softmax work onto the matrix cores and uses a high-quality, in-kernel conversion path to produce quantized probabilities.
Our Fused-Softmax pipeline combines the following key features:

\begin{itemize}
    \item Freezing the maximum speculatively, which moves the max subtraction into the score GEMM's accumulator and removes the rescaling-related vector work. A gate
    on the running tile max detects overflow and triggers a rollback, which raises the frozen
    max and re-shifts the scores.

    \item Exponential approximation using the Schraudolph method~\cite{schraudolph1999exp}, with the
    affine transformation folded into the score GEMM's operands alongside the frozen maximum, off the vector unit. 

    \item A hardware-efficient two-stage conversion to quantized $P$: \texttt{Stage-A} performs the Schraudolph conversion, and \texttt{Stage-B} scales the result to the target format. 
    The running tile max is used to derive the \texttt{E8M0} $P$ scales in the microscaling (MX) format~\cite{ocp2023mx}.

    \item The row sum is moved out of the vector path and done as an efficient matrix multiply using the quantized $P$.
\end{itemize}

This leaves only a maximum reduction and two conversions on the vector unit, effectively unlocking the
benefits of quantizing the matrix operands to FP8 and lower formats. We implement our pipeline on
CDNA~4 and evaluate it on the AMD Instinct\texttrademark{} MI355X GPU. Section~\ref{sec:ablation}
ablates the accuracy cost of each feature and the operand quantization it enables.
We evaluate a progressive series of kernel variants from MXFP8 through MXFP6 to MXFP4.
The MXFP8 configuration reaches 2978~TFLOP/s ($2.07\times$ BF16), at a cosine similarity
of 0.9984. The MXFP6 attention reaches 3973~TFLOP/s ($2.76\times$ BF16), at
0.9979. The frontier MXFP4 attention reaches 4586~TFLOP/s ($3.2\times$ BF16),
at 0.9787.

The kernels are in production: they ship in AITER behind the \texttt{mha\_v4}
entrypoint~\cite{amd2025aiter}, and
powered AMD's MLPerf\textsuperscript{\textregistered} Inference v6.1 Wan2.2-T2V-A14B
submission~\cite{amd2026mlperf61blog,amd2026mlperf61rocm}, whose single-stream latency fell
$1.7\times$ from the previous round on the same MI355X system, with quantized attention the largest
single contributor.

\section{Background}
\label{sec:background}
In this section, we establish the background of our work. We review FlashAttention and the online
softmax, introduce the asymmetric scaling of matrix and vector throughput that leaves the softmax
behind, and survey the recent work addressing it.

Throughout, uppercase symbols such as $Q$ and $S$ denote matrices, and lowercase symbols such as
$m$ and $\ell$ denote per-row quantities, which we treat as column vectors that broadcast across
the columns of a matrix wherever the two are combined. A subscript on a matrix selects a
\emph{tile} of rows rather than an entry: $Q_i$ is the $i$th block of $M$ consecutive rows of $Q$,
$K_j$ the $j$th block of $N$ rows of $K$, and a double subscript $S_{ij}$ the $M \times N$ block of
$S$ formed from $Q_i$ and $K_j$. Row-wise reductions are written $\mathrm{rowmax}$ and
$\mathrm{rowsum}$, each returning one value per row.

\subsection{FlashAttention}
\label{sec:flashattention}

Let $Q, K, V \in \mathbb{R}^{N_{\mathrm{ctx}} \times d}$ be the query, key and
value matrices of one attention head, with $N_{\mathrm{ctx}}$ the sequence
length and $d$ the head dimension. Attention computes
\begin{equation}
  S = \alpha\, Q K^{\top}, \qquad
  P = \mathrm{softmax}(S), \qquad
  O = P V,
\end{equation}
with $\alpha = 1/\sqrt{d}$ and the softmax taken row-wise.

FlashAttention \cite{dao2022flashattention,dao2023flashattention2,shah2024flashattention3}
never materializes $S$ or $P$ in device memory. A workgroup owns a single query tile $Q_i \in \mathbb{R}^{M \times d}$ of $M$
rows and loops over the keys and values $K_j, V_j \in \mathbb{R}^{N \times d}$ in tiles of $N$ rows, indexed
$j = 0,\dots,J-1$ with $J = \lceil N_{\mathrm{ctx}}/N \rceil$. Iteration $j$ computes the score
tile $S_{ij} = \alpha\, Q_i K_j^{\top}$ and folds it into the running state by online softmax
\cite{milakov2018onlinesoftmax,rabe2021selfattention}:
\begin{align}
  X_{ij} &= (\log_2 e)\,S_{ij}, \label{eq:base2score}\\
  m_j    &= \max\!\left(m_{j-1},\, \mathrm{rowmax}(X_{ij})\right), \label{eq:onlinemax}\\
  P_{ij} &= 2^{X_{ij}-m_j}, \qquad r_j = 2^{m_{j-1}-m_j}, \label{eq:onlinep}\\
  \ell_j &= r_j\,\ell_{j-1} + \mathrm{rowsum}(P_{ij}), \label{eq:onlinesum}\\
  O_j    &= r_j\,O_{j-1} + P_{ij} V_j. \label{eq:onlineout}
\end{align}
The recurrence starts from $m_{-1} = -\infty$ and $\ell_{-1} = O_{-1} = 0$, which makes the first
rescaling factor $r_0$ vanish and leaves $\ell_0$ and $O_0$ equal to the contribution of tile $0$
alone. Once the loop ends, the output accumulator is normalized by $1/\ell_{J-1}$ to complete the
softmax.

Equation~\eqref{eq:base2score} scales the score by the constant $\log_2 e$, as FlashAttention
implementations commonly do to use the hardware's base-two exponential; in this paper the choice
carries more weight. It replaces $e^{x}$ by $2^{x}$ everywhere
downstream and leaves the probabilities untouched, since $2^{X_{ij}} = e^{S_{ij}}$ exactly. Base
two is the base the hardware already works in: it is the base of the exponent field of a floating-point
number, of the hardware exponential instruction, and of the Schraudolph construction of
Section~\ref{sec:schraudolph}. Carrying the softmax in base two is what later allows the
exponential to be absorbed into a format conversion rather than evaluated.

Subtracting the row maximum before the exponential keeps the exponent and the output accumulator
inside the FP32 range, but it is also what couples the iterations: whenever the maximum grows, the
accumulated output and the running row sum computed under the old maximum have to be corrected by
$r_j$.

Each iteration therefore contains two GEMMs, $Q_iK_j^{\top}$ and $P_{ij}V_j$, separated by a body
of vector work: the row maximum, the max subtraction, the exponential, the row sum, and the
rescaling of $O$ and $\ell$ by $r_j$.

\subsection{Asymmetric scaling of matrix and vector throughput}
\label{sec:asymmetric}

Matrix and vector throughput have not been scaling the same way. As
\citet{zadouri2026flashattention4} note, successive accelerator generations have spent their power
and silicon-area budget on the matrix-multiply units, the components that dominate performance,
while the vector units have stayed close to flat. Attention runs into this directly, because the
softmax sitting between its two GEMMs is vector work and does not benefit. We quantify the gap on
CDNA, and then show that it is not a quirk of CDNA.

Table~\ref{tab:cdna4} gives matrix and vector throughput on the last two CDNA
generations, per clock per compute unit (CU), taken from the official CDNA~4 whitepaper \cite{amd2025cdna4whitepaper}.

\begin{table}[t]
\caption{Throughput evolution on the last two CDNA generations. 
Matrix, vector arithmetic logic unit (VALU), and exponential rows are ops/clock/CU;
the local data share (LDS) row is bytes/clock/CU.}
\label{tab:cdna4}
\vskip 0.1in
\begin{center}
\normalsize\setlength{\tabcolsep}{12pt}%
\begin{tabular}{lrrr}
\toprule
Unit & CDNA~3 & CDNA~4 & Gain \\
\midrule
Matrix, BF16            & 2048 & 4096  & $2\times$ \\
Matrix, FP8             & 4096 & 8192  & $2\times$ \\
Matrix, MXFP6/FP4       & ---  & 16384 & --- \\
\midrule
Vector ALU              & 128  & 128   & $1\times$ \\
Exponential unit        & 16   & 32    & $2\times$ \\
LDS read (bytes)        & 128  & 256   & $2\times$ \\
\bottomrule
\end{tabular}
\end{center}
\vskip -0.1in
\end{table}

The matrix core has gained along two independent axes, one generational and the other from the
operand format, so a kernel moving from BF16 on CDNA~3 to MXFP6 on CDNA~4 has
$8\times$ the matrix throughput. Peak VALU throughput, meanwhile, has stayed flat at $128$ elements
per clock per CU. On CDNA~4 this leaves the vector-to-matrix throughput ratios at $1{:}32$,
$1{:}64$ and $1{:}128$ for BF16, FP8 and FP6/FP4 respectively, and the exponential-to-matrix ratios
further behind still, at $1{:}128$, $1{:}256$ and $1{:}512$. Quantizing the operands therefore
widens the gap rather than closing it: the faster the format, the further the softmax falls behind.

NVIDIA hardware shows the same pattern. Derived from the published peak dense rates of the
Blackwell Ultra GPU in GB300 NVL72~\cite{nvidia2025blackwellultra, nvidia2025blackwellultradatasheet}, 
its vector-to-matrix throughput ratios are $1{:}62.5$ at BF16, $1{:}125$ at FP8 and FP6 and
$1{:}375$ at FP4, and the exponential-to-matrix ratios are about $1{:}234$, $1{:}467$ and $1{:}1402$.
For both vendors, we use the peak vector operation rate in the ratios, rather than the peak FLOP rate, because the latter only holds by counting 
a fused multiply-add as two FLOPs~\cite{amd2025cdna4whitepaper, nvidia2025cudaguide}, 
and does not generalize to the softmax vector operations of Section~\ref{sec:valuroofline}.
The two architectures make different trade-offs, but both place the vector work rate, and the
exponential rate in particular, well below the matrix rate once the operands are quantized, so the
softmax bottleneck described in this paper is not specific to CDNA~4.

\subsection{Related work addressing the softmax bottleneck}
\label{sec:relatedwork}

FlashAttention-4~\cite{zadouri2026flashattention4} attacks the exponential and the output
rescaling. Part of the exponentials is emulated on the vector units, which on NVIDIA
hardware run alongside a separate exponential unit: the integer part is evaluated exactly with the
Schraudolph construction~\cite{schraudolph1999exp}, and a third-degree polynomial approximates the
fractional part. FlashAttention-4 targets BF16 attention. 
We use the same Schraudolph construction, but with a linear fractional part: we find its error small
next to that of quantizing the attention inputs (Section~\ref{sec:ablation}), and the transformation being affine, it can be
folded into the score GEMM operands (Section~\ref{sec:schraudolph}).

For the output rescaling, FlashAttention-4 keeps a stale row maximum and rescales the output only
when the running maximum exceeds it by more than a threshold set to ensure numerical stability;
otherwise the stale row maximum is used to shift the scores and the output rescaling can be omitted.
Our frozen maximum (Section~\ref{sec:freeze}) achieves the same skipping of the output rescaling, but more importantly,
moving the frozen maximum subtraction speculatively before the score GEMM is what enables our 
affine Schraudolph transformation to be fused to the score GEMM (Section~\ref{sec:schraudolph}). The rollback scheme (Section~\ref{sec:freeze})
is what ensures the numerical stability for us.

\begin{figure}[t]
  \centering
  \includegraphics[width=\columnwidth]{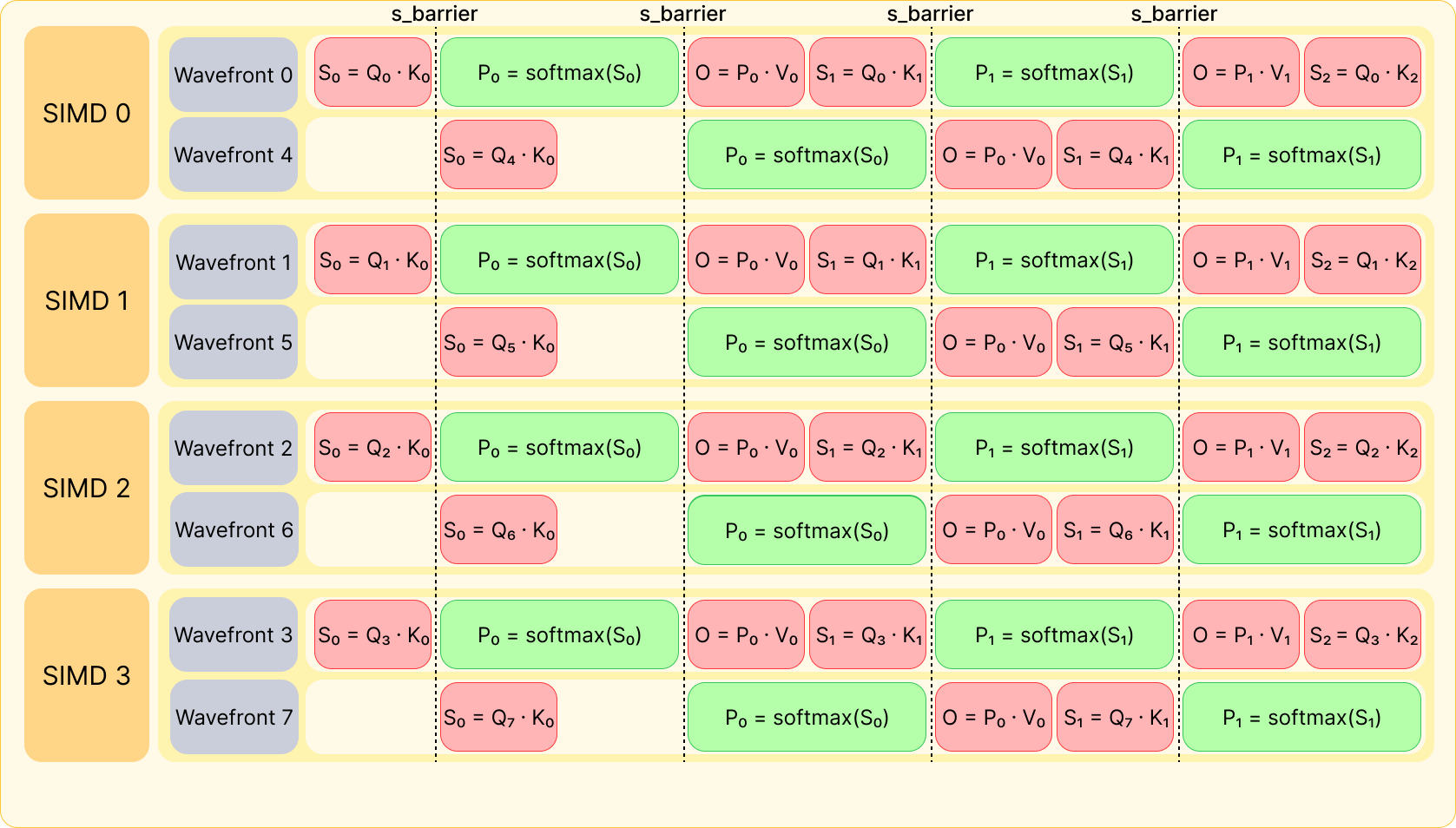}
  \caption{Eight-wave ping-pong schedule on four SIMDs. Waves $0$--$3$ and $4$--$7$ form
  phase-shifted groups: each wave overlaps its matrix operations with its SIMD partner's softmax,
  and barriers preserve the phase offset across tiles.}
  \label{fig:pingpong_scheduling}
\end{figure}

Ping-pong pipelining, introduced for attention by FlashAttention-3~\cite{shah2024flashattention3}
and kept by FlashAttention-4, phase-shifts two groups of warps so that one runs its GEMMs while
the other runs its softmax. The producer/consumer split FlashAttention-3 pairs with it does not
carry over to CDNA, where registers are allocated statically and a producer wave would hold a full
register budget while doing no arithmetic. HipKittens~\cite{hu2026hipkittens} replaces it with an
eight-wave ping-pong, which we adopt (Figure~\ref{fig:pingpong_scheduling}): a workgroup spans the
four single-instruction multiple-data (SIMD) units of a CU, waves 0--3 and 4--7 form two groups
that share a SIMD pairwise, and a barrier holds a phase offset so one group's vector instructions
issue while the other's matrix fused multiply--add (MFMA) instructions are in flight.

The SageAttention series~\cite{zhang2025sageattention,zhang2025sageattention2,zhang2025sageattention3}
quantizes the attention GEMMs progressively from INT8 to INT4 for $QK$ and FP8 to FP4 for $PV$.
Several follow-ups to FlashAttention-4 quantize its GEMMs to FP8 and below. Attn-QAT~\cite{zhang2026attnqat}
quantizes $QK$ to FP8 or FP4 and $PV$ to FP8, and recovers the accuracy lost at 4 bits by fine-tuning
the model with quantization-aware training. These approaches leave softmax unchanged. At sub-FP8
precision, softmax becomes the throughput-limiting stage in the ping-pong pipeline on modern datacenter GPUs.

Two concurrent works target the slow part of the softmax by replacing the exponential with an affine transformation.
Direct-P~\cite{hu2026fp4flashattention4} maps the scores
with a fitted linear function into MXFP4 probabilities, which relies heavily on that format having only a few
representable levels. VC-Attention~\cite{li2026vcattention} evaluates the exponential with
Schraudolph's construction~\cite{schraudolph1999exp} and reads the result as an FP8 probability in
one conversion. That single conversion to an 8-bit integer covers only a narrow range of the exponential;
Sections~\ref{sec:2stage_cvt} and~\ref{sec:ablation} find the same 8-bit proxy too costly in accuracy in our pipeline. In both
methods the affine transformation, the row maximum and the row sum stay on the vector units.

MXAttention~\cite{yu2026mxattention} targets accuracy rather than throughput. Accumulating the
output with quantized probabilities while normalizing with a higher-precision row sum biases the
result; using the quantized probabilities for both removes the bias. Our row sum, computed as a
matrix product of those probabilities, does the same (Section~\ref{sec:lsum}).

\section{Motivation: softmax bottlenecks below FP8}
\label{sec:motivation}

Here we estimate the matrix, vector, and memory work cost of FlashAttention on MI355X.
We use our implementation's actual tile sizes $M=256$, $N=128$, and $d=128$, and the Section~\ref{sec:asymmetric} rates to derive the clock cycle costs.

\subsection{GEMM roofline}
One iteration of the inner loop of FlashAttention performs two matrix multiplications,
$Q_iK_j^{\top}$ and $P_{ij}V_j$. Each matrix multiply requires $2MNd$ operations, so using the Section~\ref{sec:asymmetric} rates the two
cost
\begin{equation}
  \resizebox{0.89\columnwidth}{!}{$\displaystyle
  T^{FP8}_{\mathrm{GEMM}} = \frac{4MNd}{8192} = 2048,\quad
  T^{FP6/FP4}_{\mathrm{GEMM}} = \frac{4MNd}{16384} = 1024$}
\end{equation}
clock cycles for FP8 and FP6/FP4, respectively.
Narrowing from FP8 to FP6/FP4 halves the matrix cost.
FP6/FP4 GEMMs share a rate on CDNA~4. FP4 earns its place on memory bandwidth and
register usage reduction rather than on matrix throughput.

\subsection{VALU roofline}
\label{sec:valuroofline}
The softmax processes the $MN=32768$ score elements: it computes the row
maximum, scales and subtracts that maximum, exponentiates, sums each row, and
converts $P_{ij}$ to the format consumed by the $P_{ij}V_j$. Output rescaling acts on
$Md=32768$ accumulator elements.

Using the ideal throughput of packed instructions of $128$ elements per clock per CU
for the four passes over the score tile (max, max subtraction, row sum and $P$ conversion) and for
the output rescale, and $32$ for the exponential, we get
\begin{equation}
  T_{\mathrm{VALU}} \approx \frac{MN}{32} + 4\times\frac{MN}{128} + \frac{Md}{128} = 2304
\end{equation}
clock cycles for the vector work, of which the exponential alone accounts for $1024$ --- the entire FP6/FP4 matrix time.

\subsection{Memory bandwidth roofline}
The operands have to reach the matrix cores before they can be multiplied, over two hops:
high-bandwidth memory (HBM) into LDS, and LDS into registers. Dividing the bytes each $K_j/V_j$
tile moves by the bandwidth available per CU gives, at FP8, FP6 and FP4, about $2517$, $1887$ and
$1258$ clock cycles from HBM to LDS ($T_{\mathrm{HBM}}$) without any cache reuse, and $1024$, $768$ and $512$ clock cycles from LDS into the
registers ($T_{\mathrm{LDS}}$) (Appendix~\ref{app:memroofline} gives the derivation). Double buffering
allows HBM transfers to prefetch the next $K/V$ tile while the current one is computed, and LDS reads
are interleaved with matrix execution.

Table~\ref{tab:roofline} summarizes the per-tile HBM, LDS, matrix, and vector clock cycle costs. The vector work exceeds the matrix time
at FP8 and costs $2.25\times$ as much at FP6/FP4. HBM transfers and LDS reads are able to hide under computation.

\begin{table}[!htbp]
\caption{Per-tile roofline, in clock cycles.}
\label{tab:roofline}
\centering
\normalsize\setlength{\tabcolsep}{12pt}%
\begin{tabular}{lrrr}
\toprule
 & FP8 & FP6 & FP4 \\
\midrule
$T_{\mathrm{HBM}}$            & $2517$ & $1887$ & $1258$ \\
$T_{\mathrm{LDS}}$            & $1024$ & $768$ & $512$ \\
$T_{\mathrm{GEMM}}$           & $2048$ & $1024$ & $1024$ \\
$\bm{T_{\mathrm{VALU}}}$      & $\bm{2304}$ & $\bm{2304}$ & $\bm{2304}$ \\
\midrule
$T_{\mathrm{VALU}}/T_{\mathrm{GEMM}}$ & $1.125\times$ & $2.25\times$ & $2.25\times$ \\
\bottomrule
\end{tabular}
\end{table}

\section{Fused-Softmax Attention}
\label{sec:algorithm}

Fused-Softmax Attention removes most of the softmax's vector work from the steady-state loop by folding it
into the two GEMMs and a format conversion. The maximum is frozen speculatively and subtracted
inside the score GEMM, so the output and the row sum are never rescaled unless a later tile
overflows, in which case an exact rollback restores them. The exponential is approximated by Schraudolph's affine
map, carried by the score GEMM's MX scale and the subtraction, and so reduces to a conversion; a second conversion
quantizes $P$, and the block maxima computed along the way supply both the overflow gate and the
MX scales of $P$. The row sum leaves the vector unit as well, evaluated as a matrix multiply
over the quantized probabilities. In the steady state the vector unit runs only the block-maximum reduction
and the two conversions.

\begin{figure}[t]
  \centering
  \includegraphics[width=\textwidth]{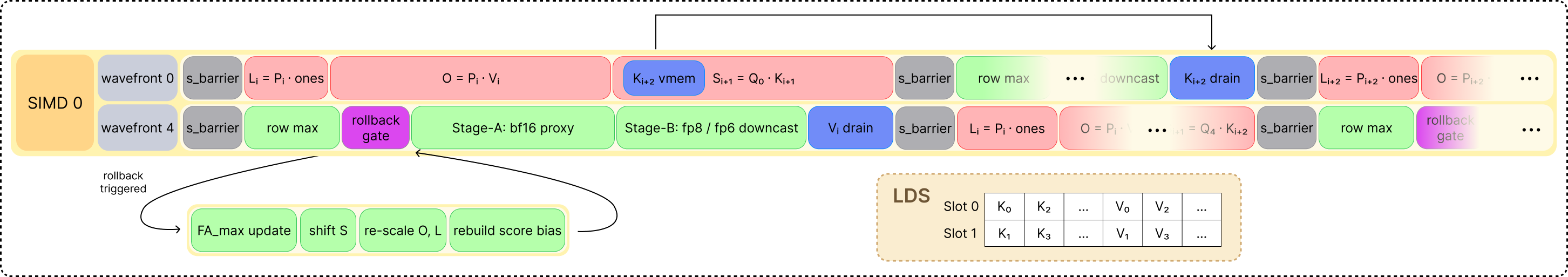}
  \caption{One SIMD pair in the Fused-Softmax schedule. While one wave issues the denominator,
  $PV$ and the next $QK$ matrix operations, its partner evaluates the row maximum, rollback gate
  and two-stage conversion. Prefetch and drain operations use alternating LDS slots; the rollback
  path below the timeline runs very rarely, only when the gate fires.}
  \label{fig:detailed_pingpong_scheduling}
\end{figure}

We adopt the eight-wave ping-pong of
Section~\ref{sec:relatedwork}, the CDNA form of the pipeline FlashAttention-3 introduced and
FlashAttention-4 kept, and fuse the softmax into it. Data movement uses the same two-group split:
the groups cooperatively prefetch $K$ and $V$ into free LDS slots while computing on the tiles
already resident in the others (Figure~\ref{fig:detailed_pingpong_scheduling}). Waves 0--3 fetch
$K$ and waves 4--7 fetch $V$, each one tile ahead into a two-slot ring, so a transfer for the next
tile is in flight for the whole of the current one, and a barrier publishes each slot before it is
read.

Algorithm~\ref{alg:fusedsoftmax} states the computation for one query tile $i$, in terms of
Eqs.~\eqref{eq:onlinemax}--\eqref{eq:onlineout}. 
$Q_i$ is preprocessed as $(\alpha\log_2 e)\,Q_i$ before quantization.
Hats denote MX elements and $s$ their \texttt{E8M0}
scales, e.g.\ $\hat Q_i$ and $s^Q_i$. $C$ and $D$ are the fused Schraudolph scales and bias, matching Section~\ref{sec:schraudolph}. Lines 1--3 and 6 fold the
Schraudolph affine transformation into the score MFMA (Section~\ref{sec:schraudolph}); lines 8--13
are the frozen maximum with rollback (Section~\ref{sec:freeze}); lines 7 and 15--16 are the two-stage
conversion of $P$, whose block maxima serve both the gate and the scales (Section~\ref{sec:2stage_cvt});
and lines 6 and 17--18 are the MX matrix multiplies, the row sum included (Section~\ref{sec:operands}).

\begin{algorithm}[t]
\small
\caption{Fused-Softmax Attention for one query tile $Q_i$. \matrixline{Matrix-core} lines are
tinted red and \vectorline{vector-unit} lines green.}
\label{alg:fusedsoftmax}
\begin{algorithmic}[1]
\STATE $C = 2^{-9},\ D = 2^{-9}(127-\epsilon)$
\STATE $m = \mathrm{rowmax}(X_{i0})$ \COMMENT{peeled tile $j=0$}
\STATE $s^Q_i \leftarrow C\,s^Q_i, \beta = D - C\,m$
\STATE $\ell_0 = P_{i0}\mathbf{1},\ O_0 = P_{i0}V_0$ \COMMENT{tile $0$, as lines 15--18}
\FOR{$j = 1, \dots, J-1$}
  \STATE \matrixline{$A_{ij} = (s^Q_i \odot \hat Q_i)(s^K_j \odot \hat K_j)^{\top} + \beta$}
  \STATE \vectorline{$b_{ij} = \mathrm{rowmax}_{32}(A_{ij})$}
  \IF[rollback]{$\mathrm{any}(b_{ij} > a_{\rm cap})$}
    \STATE \vectorline{$\Delta_j = \lceil(\mathrm{rowmax}(b_{ij}) - D)/C\rceil$}
    \STATE \vectorline{$\Delta_j \leftarrow 0$ where $\mathrm{rowmax}(b_{ij}) \le a_{\rm cap}$}
    \STATE \vectorline{$m \leftarrow m + \Delta_j,\ \beta \leftarrow D - C\,m$}
    \STATE \vectorline{$A_{ij},\, b_{ij} \leftarrow A_{ij} - C\Delta_j,\ b_{ij} - C\Delta_j$}
    \STATE \vectorline{$\ell_{j-1},\, O_{j-1} \leftarrow 2^{-\Delta_j}\ell_{j-1},\ 2^{-\Delta_j}O_{j-1}$}
  \ENDIF
  \STATE \vectorline{$\tilde P_{ij} = \mathrm{as\_bf16}(\mathrm{u16}_{\rm norm}(A_{ij})) \approx 2^{X_{ij}-m}$}
  \STATE \vectorline{$(\hat P_{ij},\ s^P_{ij}) \leftarrow \mathrm{Quant}(\tilde P_{ij},\ b_{ij})$}
  \STATE \matrixline{$\ell_j = (s^P_{ij}\odot\hat P_{ij})\,\mathbf{1} + \ell_{j-1}$}
  \STATE \matrixline{$O_j = (s^P_{ij}\odot\hat P_{ij})(s^V_j \odot \hat V_j) + O_{j-1}$}
\ENDFOR
\STATE $O_i = O_{J-1}/\ell_{J-1}$
\end{algorithmic}
\end{algorithm}

\subsection{Frozen max with rollback recovery}
\label{sec:freeze}

Rescaling the output on every tile is a large share of the softmax's vector work, and it is needed only when
the running maximum grows. A per-row speculative value is seeded from the first tile (line 2 of Algorithm~\ref{alg:fusedsoftmax}). It is frozen
and raised only when a later tile overflows (lines 8--13). This is admissible because the subtracted constant
cancels in the softmax ratio, so any value serves. The running maximum is merely the choice that
guarantees no overflow. While the speculative maximum holds, the rescaling factor $r_j$ of
Eqs.~\eqref{eq:onlinesum}--\eqref{eq:onlineout} is one, and the accumulations on lines 17--18 are
plain additions. Figure~\ref{fig:detailed_pingpong_scheduling} depicts the rollback in detail.

\paragraph{What freezing removes.}
Two of the softmax's passes over the score tile disappear with it. The rescaling reduces to the plain
addition already noted, and the max subtraction leaves the VALU entirely: a constant per row
can be folded into the score GEMM's own operands, which Section~\ref{sec:schraudolph} derives
together with the Schraudolph bias. Against the roofline of Section~\ref{sec:motivation} that is
$512$ of the $2304$\,clk the vector side would otherwise cost.

\paragraph{Only a lagging reference needs detection.}
Before rollback, the frozen reference can only lag the running maximum, so $P$ can only come out
too large and the single upper test on line 8 suffices. Recovery may conservatively place the
reference above the true maximum; this only scales subsequent probabilities downward. Rescaling the
accumulated $O$ and $\ell$ by the same power-of-two factor preserves their final ratio. The gate uses
block maxima that $P$'s \texttt{E8M0} scales already require (Section~\ref{sec:2stage_cvt}).

\paragraph{Recovery.}
Lines 9--13 carry the recovery. When a row's peak passes the gate, they shift the scores by the difference between the true and speculative maxima
(rounded up to an integer), rescale $O$ and $\ell$ by that same factor, and rebuild the
GEMM bias to match the updated maximum. The
correction is exact rather than bounded when the gate fires, so no error accumulates across tiles.

\paragraph{Threshold and cost.}
How far the maximum may lag is set by the representation of $P$ and by the range later
accumulations need. All three kernels presented here gate at $2^{32}$, that is
$a_{\rm cap} = D + 32C$ in the affine coordinate of Section~\ref{sec:schraudolph}, although the cap
is configurable. In every variant $P$ carries a power-of-two scale (Table~\ref{tab:variants}): the \texttt{E8M0} block scales of MXFP6/4 and the per-row scale of MXFP8.
The $2^{32}$ cap is a heuristic chosen for the long sequence lengths of diffusion workloads, leaving room for the subsequent accumulation in $\ell$ and $O$. Accumulation bounds also depend
on sequence length and value magnitudes; they cannot be established from
the BF16 exponent range alone.
Section~\ref{sec:rollbackrates} reports its measured trigger rates.

\subsection{Schraudolph-fused two-stage conversion to quantized P}
\label{sec:2stage_cvt}

The exponential and the narrowing of $P$ to its GEMM format are one conversion in two stages,
lines 15 and 16 of Algorithm~\ref{alg:fusedsoftmax}. \texttt{Stage-A}, line 15, builds a floating-point $P$
by reinterpreting an integer bit pattern; \texttt{Stage-B}, line 16, scales and converts that value to the microscaled
format the $PV$ GEMM consumes.

\paragraph{Schraudolph approximation.}
Schraudolph's observation \cite{schraudolph1999exp} is that a float stores a biased exponent and a
mantissa in consecutive fields, so a power of two can be built as a bit pattern instead of
computed. For a score $S$ in natural-log units,
\begin{equation}
  \resizebox{0.88\columnwidth}{!}{$\displaystyle
    P \approx \texttt{read\_as\_float}
        \bigl((C_{\rm bits} \cdot S + D_{\rm bits}).\texttt{to(uint)}\bigr)
  $}
  \label{eq:schraudolph}
\end{equation}
where a target format of $\mathrm{M}$ mantissa bits and $\mathrm{E}$ exponent bits, of bias
$b=2^{\mathrm{E}-1}-1$, takes $C_{\rm bits}=2^{\mathrm{M}}\log_2 e$ and
$D_{\rm bits}=2^{\mathrm{M}}(b-\epsilon)$. The exponent field receives the integer part of
$S\log_2 e$ and the mantissa field the fractional part, interpolated linearly over each doubling
interval,\footnotemark{} while the true function varies exponentially;
$\epsilon\approx0.05798$ is a tuned value that minimizes the root-mean-square (RMS) interpolation
error. The exponential is therefore one FMA and one cast to an unsigned integer whose bits are read
as the target floating-point format; this cast is \texttt{Stage-A}.

\footnotetext{We use \emph{doubling} for a factor-of-two interval, the span from $2^{k}$ to
$2^{k+1}$, which is one step of the binary exponent.}

\begin{table}[t]
  \caption{The three operand configurations. Throughout our kernels FP8 operands are E4M3,
  FP6 operands are E2M3 and FP4 operands are E2M1. Per-row FP8 $P$ carries a power-of-two
  scale per row and key tile.}
  \label{tab:variants}
  \centering
    \normalsize\setlength{\tabcolsep}{6pt}%
\resizebox{\linewidth}{!}{%
\begin{tabular}{lcccc}
  \toprule
  kernel & $Q$ & $K$ & $P$ & $V$ \\
  \midrule
  \texttt{mxfp8} & MXFP8 & MXFP8 & per-row FP8 & per-tensor FP8 \\
  \texttt{mxfp6} & MXFP6 & MXFP6 & MXFP6 & MXFP6 \\
  \texttt{mxfp4} & MXFP4 & MXFP4 & MXFP6 & MXFP4 \\
  \bottomrule
  \end{tabular}}
\end{table}

\paragraph{Choosing the proxy width.}
$QK$ leaves the scores in FP32, and \texttt{Stage-A} can cast them to an unsigned integer of any width,
the \textbf{Schraudolph proxy}, whose bits are then read as a float of the same width and narrowed by
\texttt{Stage-B}. There are three options:
\begin{itemize}
  \item \textbf{U32}, read as FP32 and then scaled to the $P$ format. This is the naive route, but the
  32-bit conversion has no packed form, so it misses the packed conversion family and is slow.
  \item \textbf{U16}, read as BF16 and then scaled to the $P$ format. Both stages have packed
  instructions, so this route is fast. BF16 keeps the eight exponent bits of FP32, so the dynamic range
  of $P$ available to the rollback gate is unchanged, and its seven mantissa bits give up little
  accuracy.
  \item \textbf{U8}, read directly as E4M3, in a single stage. It has no packed form either, so it is
  no faster than the two-stage U16 route, and its three mantissa bits and four exponent bits cost too
  much accuracy and range.
\end{itemize}

The accuracy cost of the U16 proxy is small. The Schraudolph approximation has an intrinsic $1.770\%$ RMS
relative error that no width removes, and U16 ($\mathrm{M}=7$) raises that only to $1.777\%$, whereas U8 ($\mathrm{M}=3$) raises the
RMS error to $3.141\%$.
Section~\ref{sec:ablation} measures the effect of each proxy on the attention output, and
Appendix~\ref{app:proxywidth} quantifies the accuracy of each proxy width.

We therefore adopt the U16 proxy in all three kernels, and \texttt{Stage-B} scales and converts the BF16 value into the $P$ format of Table~\ref{tab:variants}.

\paragraph{Reusing the maximum.}
The row maximum of line 7 is computed once and used twice: its per-block intermediate results become
the \texttt{E8M0} scales of the microscaled $P$, and the rollback gate tests the same values, so the
gate needs no reduction of its own.

\paragraph{Placing the reduction.}
Algorithm~\ref{alg:fusedsoftmax} shows the default ordering: line 7 reduces the FP32 affine
coordinates before \texttt{Stage-A}. The reduction can instead follow \texttt{Stage-A} and operate on
the packed 16-bit codes, which is more efficient. The two orders differ in how far above the frozen
maximum a score can be and still be recovered.
The FP32 coordinates represent any score the $QK$ GEMM can produce, whereas the 16-bit code saturates
$385$ doublings above the frozen maximum, beyond which the gate still fires, but the true overshoot is
lost. Our MXFP6 kernel uses the exact unsigned packed reduction throughout. MXFP4 uses the faster
\texttt{maximum3} reduction in the steady state and falls back to the unsigned reduction during
rollback. Appendix~\ref{app:f16max} scopes its guaranteed range and empirical behavior.

\subsection{Affine Schraudolph transformation and speculative max subtraction in the MX MFMA}
\label{sec:schraudolph}

Lines 1--3 and 6 of Algorithm~\ref{alg:fusedsoftmax} are the FMA of
Eq.~\eqref{eq:schraudolph}, moved off the vector unit and into the score GEMM. The cast is
irreducible and becomes \texttt{Stage-A} of Section~\ref{sec:2stage_cvt}; both of the
multiply--add's constants, and the speculative maximum, are known before that GEMM runs, so its
operands carry them.

For the U16 proxy read as BF16, $\mathrm{M}=7$ and $\mathrm{E}=8$, so $b=127$. With $X$ the base-two score of
Eq.~\eqref{eq:base2score} and $m$ the speculative score maximum, the softmax scale and
the change of logarithm base are folded into the preparation of the QK operands, which is why no
$\log_2 e$ survives in the constants below. \texttt{Stage-A} must then construct the BF16 pattern
\[
\begin{gathered}
 \text{pattern}=C_{\rm bits}(X-m)+D_{\rm bits},\\
 C_{\rm bits}=2^{7},\qquad D_{\rm bits}=2^{7}(127-\epsilon),
\end{gathered}
\]
A doubling corresponds to $2^{7}=128$ pattern units.

The conversion does not accept that pattern directly. The only available FP32-to-U16 conversion on CDNA~4 is a packed normalized conversion that multiplies
its clamped input by $65535$ and rounds to an unsigned 16-bit code, so what the GEMM must deliver
is the pattern scaled into $[0,1]$. Dividing through by $2^{16}$ gives the affine coordinate
\[
\begin{gathered}
 a=C(X-m)+D=C\,X+\beta,\\
 C=2^{-9},\qquad D=2^{-9}(127-\epsilon),\qquad \beta=D-C\,m,
\end{gathered}
\]
and $65535a$ then approximates the pattern above, the factor $65535/65536$ accounting for the
small normalization difference.

The additional factor $C$ is a power of two, so it can be folded into the
resident $Q$ MX exponents by subtracting nine. The speculative reference
contributes the per-row bias $\beta$. A BF16 MFMA broadcasts this bias into
the accumulator seed, and the microscaled QK MFMA adds the scaled product.
Its output is therefore $a$ directly. Bias preparation is repeated when the
rollback triggers; the steady-state score tile needs no separate VALU
affine transformation.

\subsection{The GEMMs: operands and instructions}
\label{sec:operands}

We quantize the $QK$ and $PV$ GEMMs progressively from FP8 through FP6 to FP4 (Table~\ref{tab:variants}). Besides these two GEMMs, the denominator accumulation moves to the matrix cores as well.

\paragraph{The QK and PV GEMMs.}
\label{sec:qkpv}
The operands of lines 6 and 18 are quantized as in Table~\ref{tab:variants}. MXFP6 and MXFP4 lower
$P$ through the same block-scaled FP6 path while retaining MXFP6 and MXFP4 $V$, so their $PV$ GEMMs
mix FP6$\times$FP6 and FP6$\times$FP4, respectively. MXFP8 lowers $P$ to per-row scaled FP8.
Before quantization, $Q$ and $K$ are rotated by a normalized Walsh--Hadamard transform $H$ over the head
dimension. Since $HH^\top = I$, $(QH)(KH)^\top = QK^\top$ and the scores are unchanged, while every rotated
element mixes all $d$ elements of its row, spreading outliers across the 32-element blocks that share an
\texttt{E8M0} scale. This processing comes from weight quantization~\cite{chee2023quip,tseng2024quipsharp}
and was brought to attention by FlashAttention-3~\cite{shah2024flashattention3}; we use a fixed (unrandomized)
Hadamard transform, fused with the MX quantization of $Q$ and $K$ into a single pass over the tensors before the attention kernel. 

\paragraph{The denominator as a GEMM.}
\label{sec:lsum}
Line 17 evaluates the softmax denominator $\ell = \sum_j P_{ij}\mathbf{1}$, a row reduction that on the VALU would compete for
exactly the issue slots the softmax already saturates. It is instead a third matrix
multiply, $P$ against a column of ones, which moves it onto a unit the softmax phase leaves idle.
The format of $P$ then decides the throughput of this matrix multiply. For an FP8 $P$ we have the
sparse FP8 matrix instruction available, which runs at twice the dense FP8 rate, the same rate as the MXFP6 and MXFP4 matrix multiplies, 
which serve the FP6 $P$ of those kernels. Either way the denominator leaves the VALU.

\section{Empirical evaluation}
\label{sec:eval}

We evaluate in two settings. The first uses \texttt{normal} inputs (a standard normal
distribution), on which the related methods Attn-QAT~\cite{zhang2026attnqat} and Direct-P~\cite{hu2026fp4flashattention4} report their accuracy: we ablate the components of the
Fused-Softmax pipeline (Section~\ref{sec:ablation}) and compare accuracy and throughput against these two
methods (Section~\ref{sec:relatedcomp}). The second uses real data captured per attention call from
Wan2.2-T2V-A14B~\cite{wan2025wan} on the MLPerf\textsuperscript{\textregistered} Inference text-to-video
benchmark~\cite{reddi2020mlperf,mlcommons2026inference} (Section~\ref{sec:diffusion}): we measure the effect of
rollback on accuracy and throughput, then analyze the model output for visual quality, score it with
VBench~\cite{huang2024vbench} and measure the end-to-end speedup.
Quality runs keep all non-attention model operations in BF16 to isolate the effect of attention
quantization. For latency, we also pair our kernels with an optimized model configuration that uses
quantized GEMMs and parallel variational autoencoder (VAE) decoding.

\subsection{Component ablation}
\label{sec:ablation}

We enable the components of the Fused-Softmax pipeline one at a time (Table~\ref{tab:ablation}), first on
BF16 attention, where $Q$, $K$, $V$ and $P$ are BF16 and the components are the only source of error beyond BF16
rounding, and then on the three quantized kernels. Starting from the conventional softmax on each format's
operands, we add the row sum as a GEMM over the quantized $P$, the frozen maximum and the Schraudolph
exponential, and finally vary the Schraudolph proxy of the complete pipeline. All configurations run on ten
seeds of the \texttt{normal} input at $b{=}1$, $h{=}24$, $s_q{=}s_k{=}32768$, $d{=}128$; we report the global
relative error $\lVert O-O_{\rm ref}\rVert_F/\lVert O_{\rm ref}\rVert_F$ and the cosine similarity against
BF16 attention with the conventional softmax, averaged over the seeds.
The complete pipeline on the quantized formats is measured on our kernels. The other configurations exist
only as steps of the ablation, so we evaluate them in a PyTorch emulator of the kernels, whose output for the
complete pipeline matches that of the kernels to the precision of Table~\ref{tab:ablation}.

\begin{table}[t]
\caption{Component ablation on \texttt{normal} inputs ($b{=}1$, $h{=}24$, $s{=}32768$, $d{=}128$, ten seeds):
global relative error in \% and cosine similarity against BF16 attention with the conventional softmax. Each
row enables the marked components. The
bolded row is our complete Fused-Softmax pipeline; the last two rows vary the Schraudolph proxy of the complete pipeline.}
\label{tab:ablation}
\centering
\small\setlength{\tabcolsep}{4pt}%
\resizebox{\linewidth}{!}{%
\begin{tabular}{@{}ccc c rr rr rr rr@{}}
\toprule
\multicolumn{4}{c}{components} & \multicolumn{2}{c}{BF16} & \multicolumn{2}{c}{MXFP8} & \multicolumn{2}{c}{MXFP6} & \multicolumn{2}{c}{MXFP4} \\
\cmidrule(r){1-4}\cmidrule(lr){5-6}\cmidrule(lr){7-8}\cmidrule(lr){9-10}\cmidrule(l){11-12}
$\ell$ as GEMM & frozen max & Schraudolph & proxy & err & cos & err & cos & err & cos & err & cos \\
\midrule
           &            &            & --  & --    & --      & 5.39 & 0.99855 & 6.17 & 0.99810 & 20.45 & 0.97890 \\
\checkmark &            &            & --  & 0.026 & 1.00000 & 5.39 & 0.99855 & 6.17 & 0.99810 & 20.45 & 0.97890 \\
\checkmark & \checkmark &            & --  & 0.310 & 1.00000 & 5.42 & 0.99853 & 6.17 & 0.99810 & 20.45 & 0.97890 \\
\checkmark & \checkmark & \checkmark & \textbf{U16} & \textbf{1.798} & \textbf{0.99984} & \textbf{5.69} & \textbf{0.99838} & \textbf{6.46} & \textbf{0.99791} & \textbf{20.53} & \textbf{0.97872} \\
\midrule
\checkmark & \checkmark & \checkmark & U32 & 1.798 & 0.99984 & 5.69 & 0.99838 & 6.46 & 0.99791 & 20.53 & 0.97872 \\
\checkmark & \checkmark & \checkmark & U8  & 3.69  & 0.99934 & 5.99 & 0.99822 & 6.71 & 0.99775 & 20.64 & 0.97854 \\
\bottomrule
\end{tabular}}
\vskip -0.1in
\end{table}

Freezing the maximum and computing the row sum as a GEMM over the quantized $P$ cost almost no accuracy. On BF16 attention, where
nothing else perturbs the output, the two together move it by at most $0.31\%$. The \texttt{normal} input never
triggers the rollback gate, so this tests the frozen maximum alone; Section~\ref{sec:rollbackrates} evaluates the
rollback on real model data.

The Schraudolph exponential is the one component with a measurable cost.
It brings BF16 attention to $1.8\%$ error, matching the approximation's own error per probability.

\textbf{Once the operands are quantized, the quantization error dominates the Fused-Softmax pipeline's accuracy cost.} On every
quantized kernel the complete pipeline adds at most $0.3$ percentage points to the error that quantizing the
operands already causes: about $5\%$ of that error at MXFP8 and MXFP6, and under $1\%$ at MXFP4.

\textbf{The U16 proxy is as accurate as the full-width U32.} U16 matches U32 on every
kernel; U8, however, doubles the BF16 error and adds error on every quantized kernel.
This further supports choosing U16 over U8 as the proxy (Section~\ref{sec:2stage_cvt}).

\begin{table}[t]
\caption{Throughput and BF16-reference cosine on \texttt{normal} input
($b{=}1$, $h{=}24$, $s{=}32768$, $d{=}128$): our kernels versus published
Attn-QAT~\cite{zhang2026attnqat} and Direct-P~\cite{hu2026fp4flashattention4} results.}
\label{tab:relatedcomp}
\centering
\normalsize\setlength{\tabcolsep}{5pt}%
\resizebox{\linewidth}{!}{%
\begin{tabular}{@{}l l rr l rr@{}}
\toprule
 & \multicolumn{3}{c}{ours} & \multicolumn{3}{c}{related work} \\
\cmidrule(lr){2-4}\cmidrule(lr){5-7}
tier & kernel & TFLOP/s & cosine & method & TFLOP/s & cosine \\
\midrule
FP8     & \texttt{mxfp8} & $2978$ & $0.99838$ & Attn-QAT MXFP8/FP8 & $2362$ & $0.9985$ \\
$\sim$FP6 & \texttt{mxfp6} & $3973$ & $0.99791$ & Attn-QAT NVFP4/FP8 & $2677$ & $0.9899$ \\
FP4     & \texttt{mxfp4} & $4586$ & $0.97872$ & Direct-P           & $2945$ & $0.9429$ \\
\bottomrule
\end{tabular}}
\vskip -0.1in
\end{table}

\subsection{Comparison with related work}
\label{sec:relatedcomp}
Table~\ref{tab:relatedcomp} places our kernels next to the throughput and cosine published for the related
methods of Section~\ref{sec:background}, one per operand
tier. We measure our kernels on the \texttt{normal} input at $b{=}1$, $h{=}24$, $s{=}32768$,
$d{=}128$, non-causal forward, the shape at which Attn-QAT and Direct-P publish their throughput,
and the shape also used in the ablation study. Our cosines are those of the complete pipeline of Table~\ref{tab:ablation}, on the
\texttt{normal} input, averaged over ten seeds.
For the related papers we report their published numbers on B300-class hardware (NVIDIA
B300/GB300) at this shape; Attn-QAT publishes its throughput on GB300 only, and its per-call cosines on B200.

VC-Attention~\cite{li2026vcattention} publishes no absolute throughput, only a $1.47\times$ speedup
over BF16 FlashAttention-4 on B300. Deriving its throughput from that claim and the BF16 FlashAttention-4
throughput Attn-QAT reports on GB300 at our shape ($1533$~TFLOP/s) gives about $2254$~TFLOP/s,
which is in line with the Attn-QAT MXFP8/FP8 throughput ($2362$~TFLOP/s).

\begin{figure}[t]
  \centering
  \includegraphics[width=\textwidth]{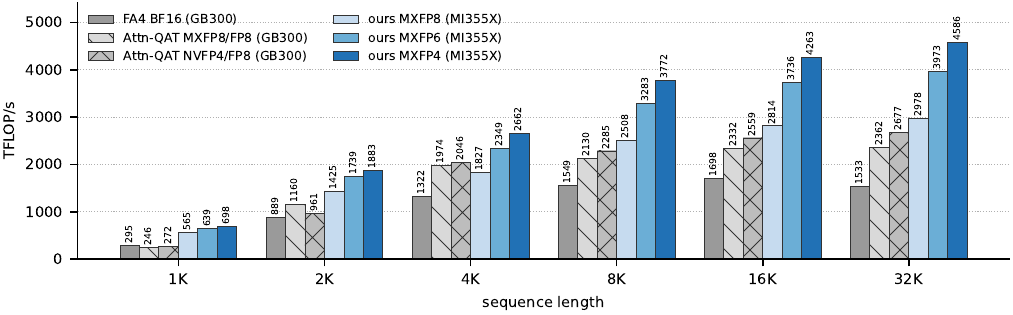}
  \caption{Forward throughput against sequence length at $b{=}1$, $h{=}24$, $d{=}128$, non-causal: our
  kernels on MI355X next to the BF16 FlashAttention-4 and Attn-QAT throughput published on
  GB300~\cite{zhang2026attnqat}.}
  \label{fig:seqlen}
\end{figure}

\textbf{Our kernels are faster than the published related work at matching or better accuracy.}
From Table~\ref{tab:relatedcomp} we can see that our kernels' cosines match or exceed the published cosines of the related work, while being
$1.26$--$1.56\times$ faster. Figure~\ref{fig:seqlen} extends the throughput comparison to the sequence
lengths for which Attn-QAT publishes its GB300 throughput, $1$K to $32$K tokens at $b{=}1$, $h{=}24$,
$d{=}128$, non-causal, where this trend continues. Our kernels are measured as above; the published
bars are read off Attn-QAT's figure.

\subsection{Video generation with Wan2.2}
\label{sec:diffusion}
We evaluate the kernels in Wan2.2-T2V-A14B~\cite{wan2025wan} with the settings of the
MLPerf\textsuperscript{\textregistered} Inference text-to-video benchmark\footnote{The text-to-video
benchmarking methodology is based on the reference implementation for Wan-2.2-T2V-A14B within the
MLPerf\textsuperscript{\textregistered} Inference Datacenter v6.1
suite~\cite{reddi2020mlperf,mlcommons2026inference}. MLPerf\textsuperscript{\textregistered} is a
registered trademark of the MLCommons\textsuperscript{\textregistered} Association. Results not verified
by MLCommons Association.}: its $248$ prompts from VBench~\cite{huang2024vbench}, 720p videos of $81$
frames generated in $20$ denoising steps from a fixed initial latent, and its six-dimension VBench
score. The one change is the scheduler's flow shift, set to Wan2.2's own value of $12$, because at
the reference's default of $3$ the videos degrade even with BF16 attention (Appendix~\ref{app:vbench}
shows this and gives the rest of the setup). All quality runs use the
model with BF16 GEMMs and BF16 cross-attention, and only the self-attention kernel changes between them. Besides the three kernels we evaluate a hybrid
schedule that runs MXFP6 for the first and last three of the $20$ denoising steps and MXFP4 in between.

On two example prompts, ``a motorcycle cruising along a coastal highway'' and ``a cat grooming itself
meticulously with its tongue'', we analyze the rollback on every self-attention call and the visual
quality of the videos, and on the motorcycle prompt the end-to-end latency. Over all $248$ prompts of
the benchmark we report the VBench scores.

\subsubsection{Per-call analysis of the rollback}
\label{sec:rollbackrates}
The \texttt{normal} input never triggers the rollback gate, so the ablation study above tests only the
frozen maximum. To test the rollback we capture every self-attention call of the BF16 generations of
the two example prompts (Figure~\ref{fig:wan22frames}; Appendix~\ref{app:vbench}). For each call we measure
the rollback rate, the fraction of key tiles in which any of the workgroup's eight waves rolls back at
the $2^{32}$ gate. Rollback is rare: over the $3200$ calls it occurs on $0.063\%$ of workgroup tiles on
average and on at most $0.67\%$ in the worst call.

\textbf{Rollback costs no measurable throughput or accuracy, and it recovers accurately.}
We take the five calls with the highest rollback rates and run our kernels on them with the gate enabled and
disabled. Even on these calls the rollback leaves the throughput unchanged within run-to-run variation, and it
costs no accuracy: against the model's BF16 attention output, our kernels are as accurate as the conventional
softmax on the same quantized operands, in every format within $0.1$ percentage points
(Appendix~\ref{app:vbench} gives the numbers). With the gate disabled every kernel's output is NaN: the frozen
maximum alone overflows on these inputs, so the rollback is needed.

\subsubsection{Visual quality}
\label{sec:visual}
Figure~\ref{fig:wan22frames} shows the same frame of both prompts through each attention configuration.
MXFP8 and MXFP6 are practically indistinguishable from the BF16 baseline. MXFP4 drifts: the cat loses its tongue and
softens, and the biker's road and foreground change. Such artifacts of FP4 attention in diffusion models
have been noted and addressed in earlier work~\cite{zhang2024sageattention2tr,zhang2026attnqat}; the hybrid
schedule removes them and stays close to the baseline.

\subsubsection{End-to-end latency}
\label{sec:e2e}

We measure the end-to-end latency on a single MI355X GPU. After one warm-up run, we generate
the motorcycle prompt five times and report the slowest run (Table~\ref{tab:vbenchmain}). The
self-attention speedup of our kernels carries over to video generation, and the hybrid schedule keeps
nearly all of MXFP4's speedup. With the model GEMMs quantized to FP8/FP4 and VAE decoding
parallelized, attention accounts for a larger share of the latency and the speedups grow further.

\subsubsection{VBench scores}
\label{sec:vbench}
We generate all 248 prompts with each attention configuration and score the videos with
VBench~\cite{huang2024vbench} on the six dimensions of the MLPerf evaluation (Table~\ref{tab:vbenchmain}; per
dimension in Appendix~\ref{app:vbench}). MXFP8, MXFP6 and the hybrid schedule match the BF16 score; only MXFP4
on its own falls noticeably behind.
VBench scores each video on its own (temporal consistency, motion, match to the prompt) rather than by its
closeness to the baseline. Since every configuration starts from the same initial latent, we also measure each
video's fidelity to the BF16 video of the same prompt, on every fourth frame, with peak signal-to-noise
ratio (PSNR), structural similarity (SSIM) and learned perceptual image patch similarity
(LPIPS). These order
the configurations by precision: MXFP8 stays closest to the BF16 generation, MXFP6 next and MXFP4 furthest,
while the hybrid schedule comes close to MXFP6's fidelity at nearly MXFP4's speed.

\begin{table}[t]
\caption{Wan2.2-T2V-A14B: VBench score and PSNR (dB), SSIM and LPIPS against the BF16 videos over
the 248 prompts. The last two columns report end-to-end latency (s) of the motorcycle prompt under
the BF16 and optimized model configurations of Section~\ref{sec:e2e}; parentheses give the speedup
over BF16 attention in the same configuration.}
\label{tab:vbenchmain}
\centering
\normalsize\setlength{\tabcolsep}{5pt}%
\resizebox{\linewidth}{!}{%
\begin{tabular}{lrrrr|rr}
\toprule
 & & & & & \multicolumn{2}{c}{latency (s)} \\
 & VBench & PSNR $\uparrow$ & SSIM $\uparrow$ & LPIPS $\downarrow$ & BF16 GEMMs & FP8/4 GEMMs \\
\midrule
BF16   & 71.83 & \multicolumn{3}{c|}{reference} & 197.5 & 173.3 \\
MXFP8  & 71.78 & 17.37 & 0.584 & 0.324 & 129.4 ($1.53\times$) & 105.9 ($1.64\times$) \\
MXFP6  & 71.85 & 16.32 & 0.546 & 0.363 & 115.5 ($1.71\times$) &  88.9 ($1.95\times$) \\
MXFP4  & 71.13 & 14.55 & 0.477 & 0.486 & 106.0 ($1.86\times$) &  82.6 ($2.10\times$) \\
hybrid & 72.01 & 16.31 & 0.540 & 0.382 & 108.0 ($1.83\times$) &  84.4 ($2.05\times$) \\
\bottomrule
\end{tabular}}
\vskip -0.1in
\end{table}

\begin{figure}[t]
  \centering
  \setlength{\gridw}{\dimexpr(\textwidth-\gridlab)/5\relax}
  {\offinterlineskip
  \hbox{\makebox[\gridlab]{}\gridhead{BF16}\gridhead{MXFP8}\gridhead{MXFP6}\gridhead{MXFP4}\gridhead{hybrid (MXFP6 + MXFP4)}}\vspace{1pt}
  \hbox{\gridrow{cat}\gridimg{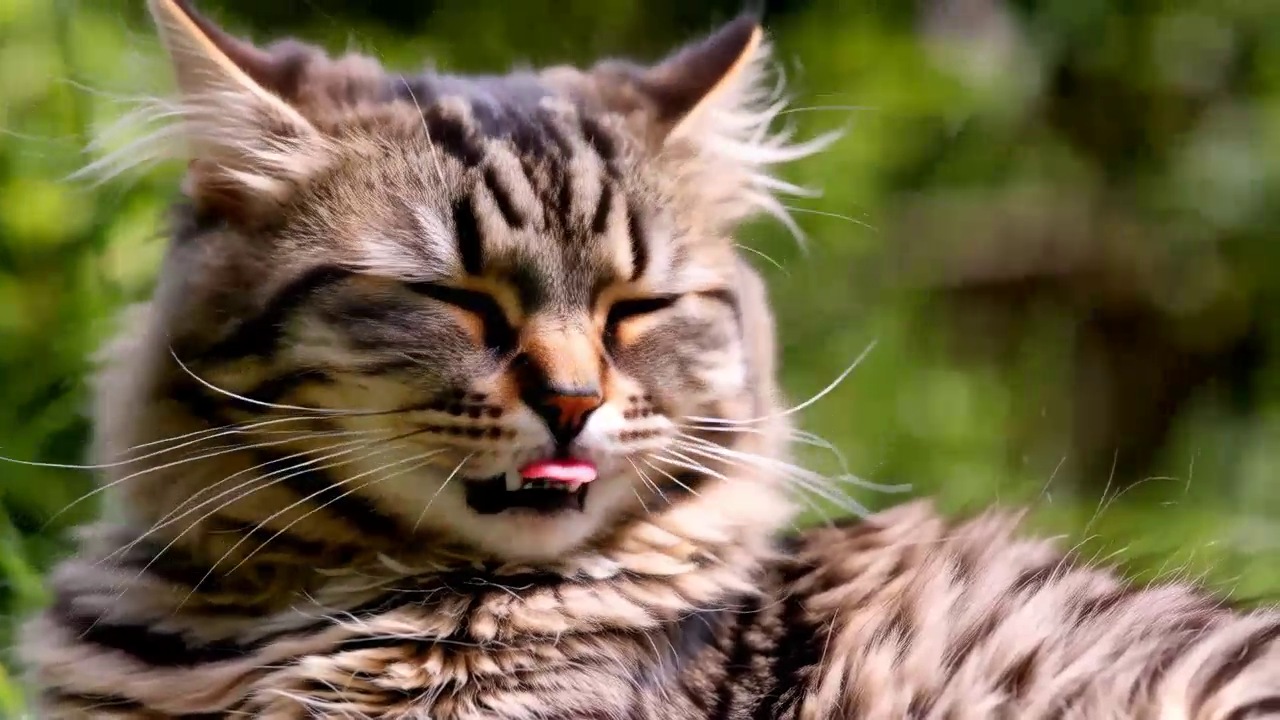}\gridimg{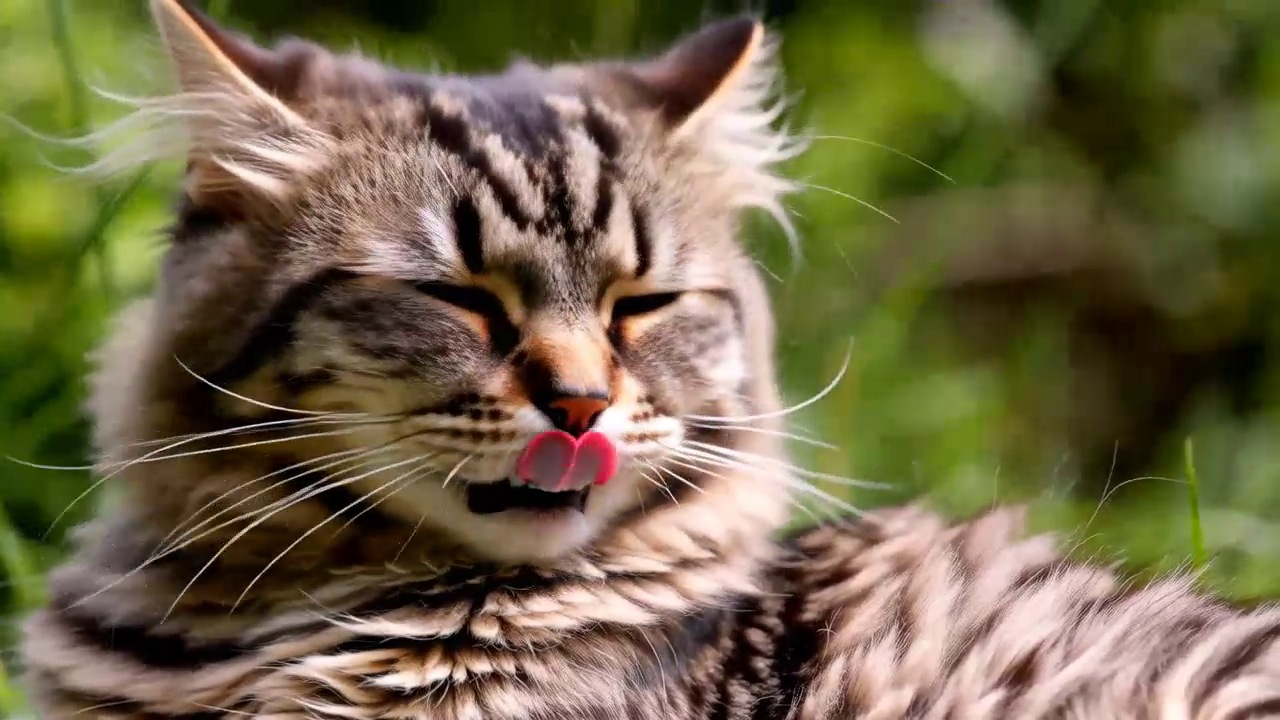}\gridimg{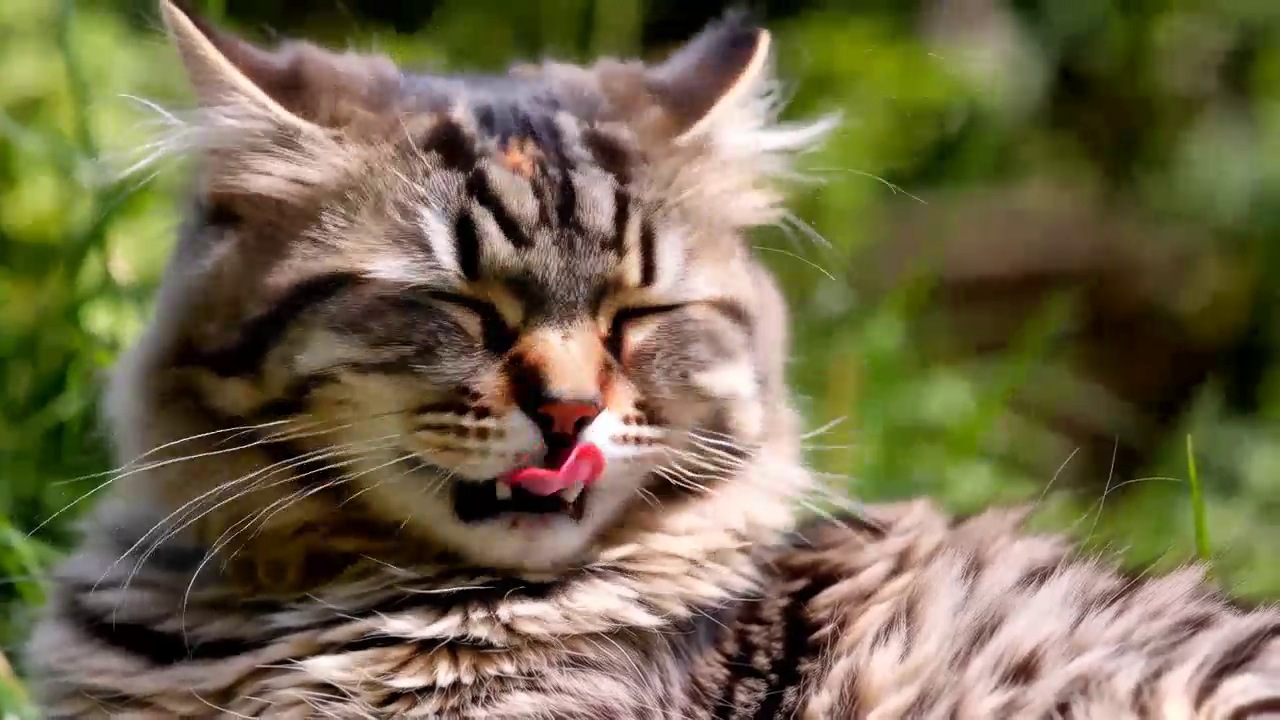}\gridimg{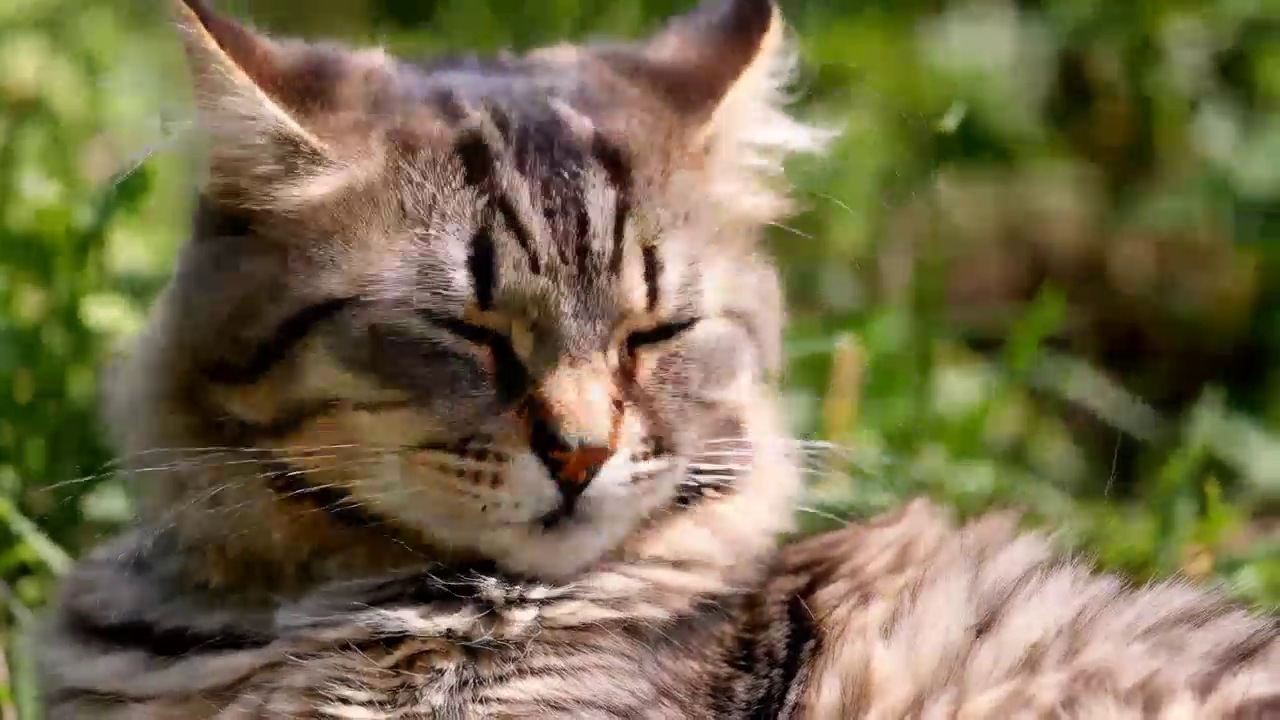}\gridimg{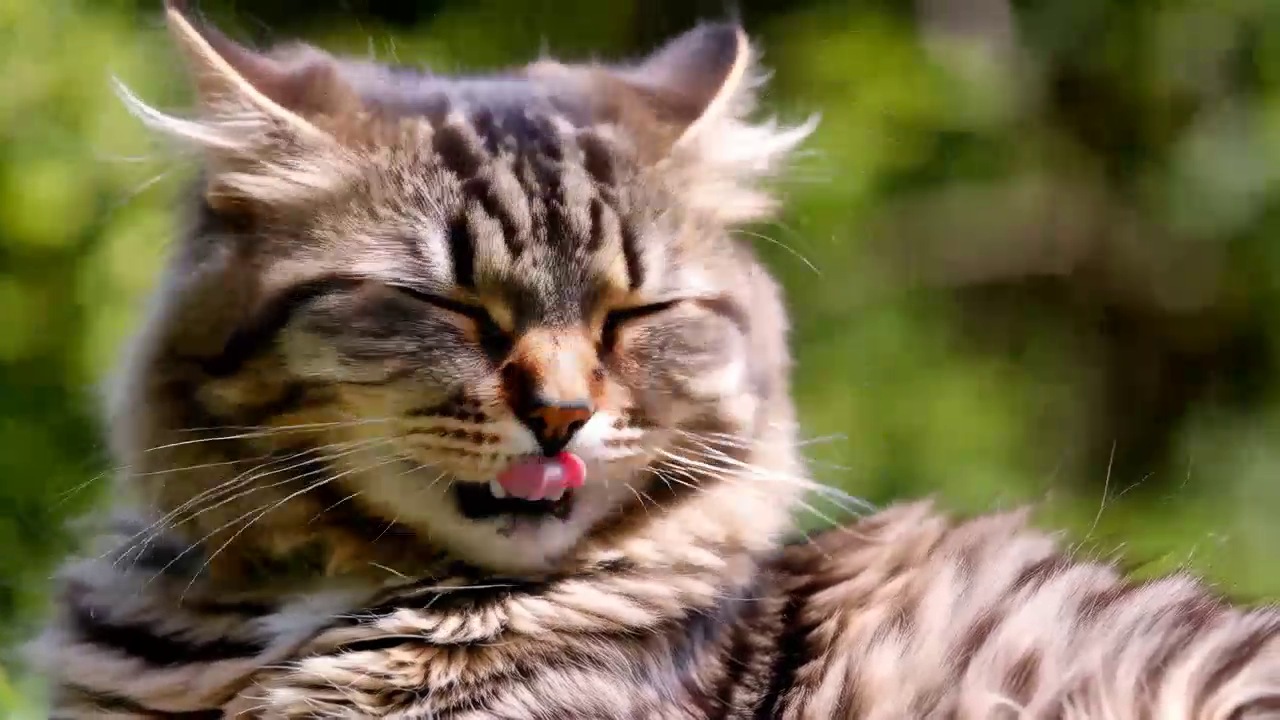}}
  \hbox{\gridrow{motorcycle}\gridimg{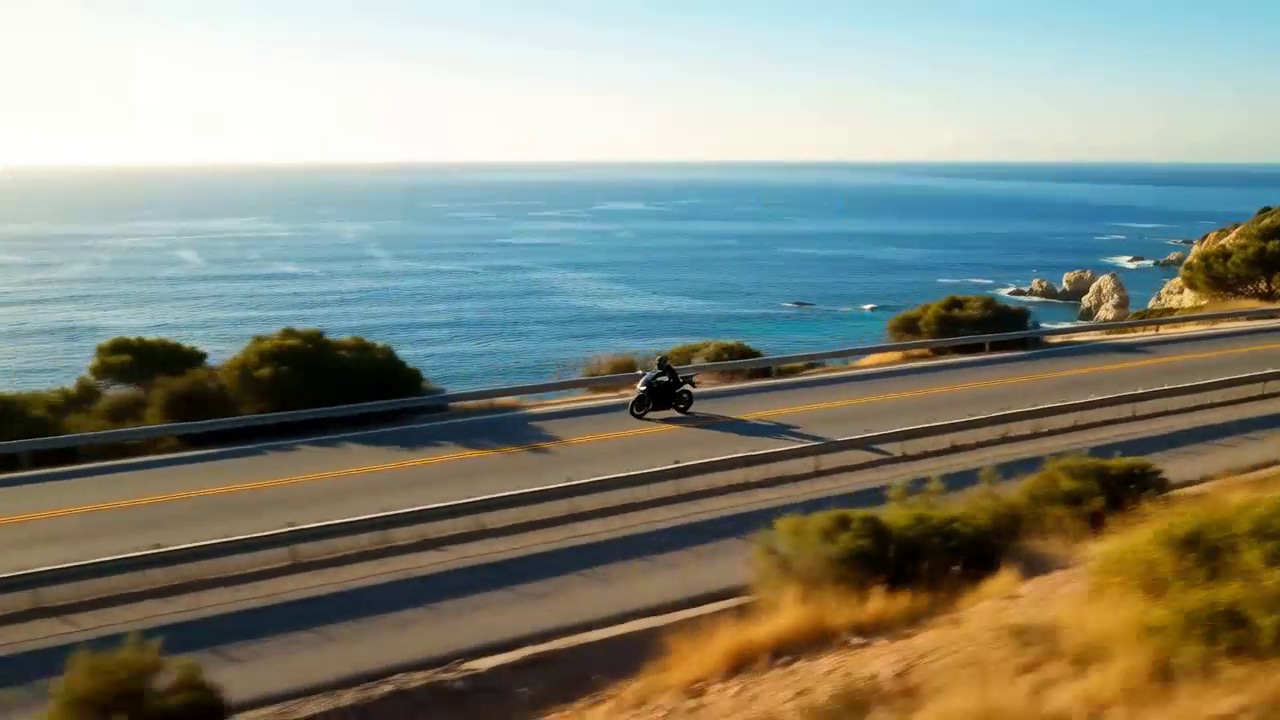}\gridimg{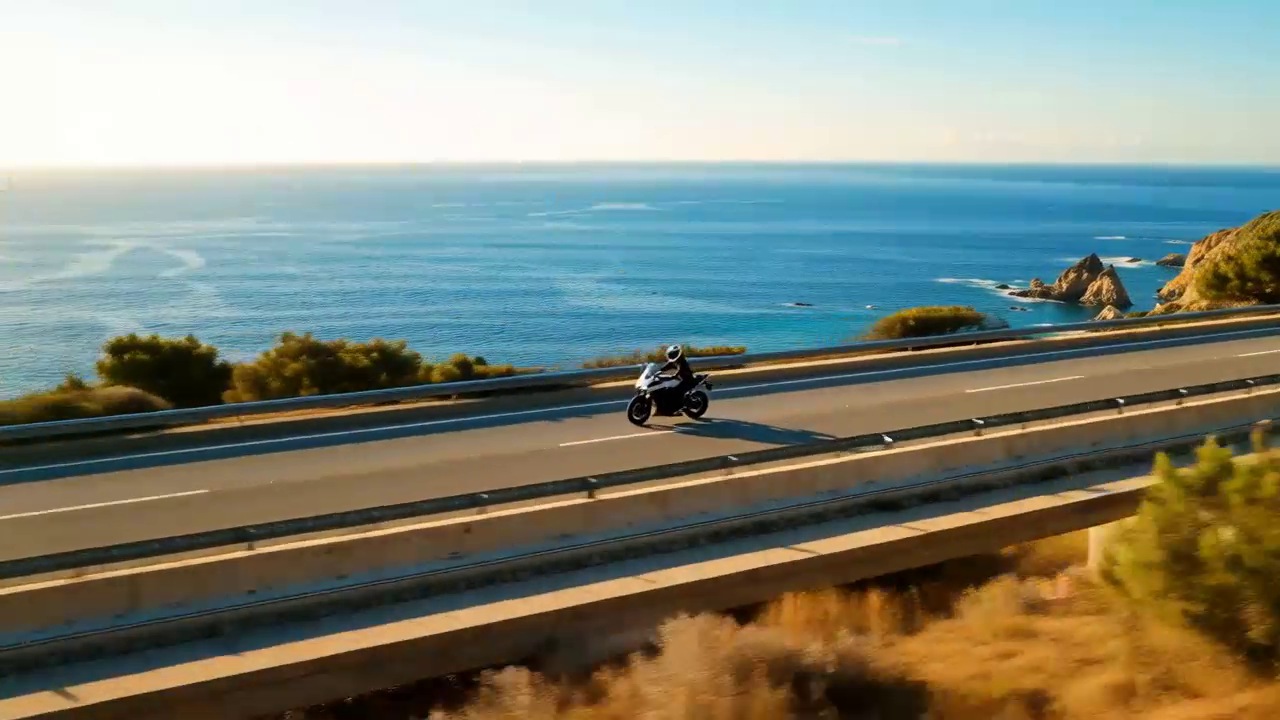}\gridimg{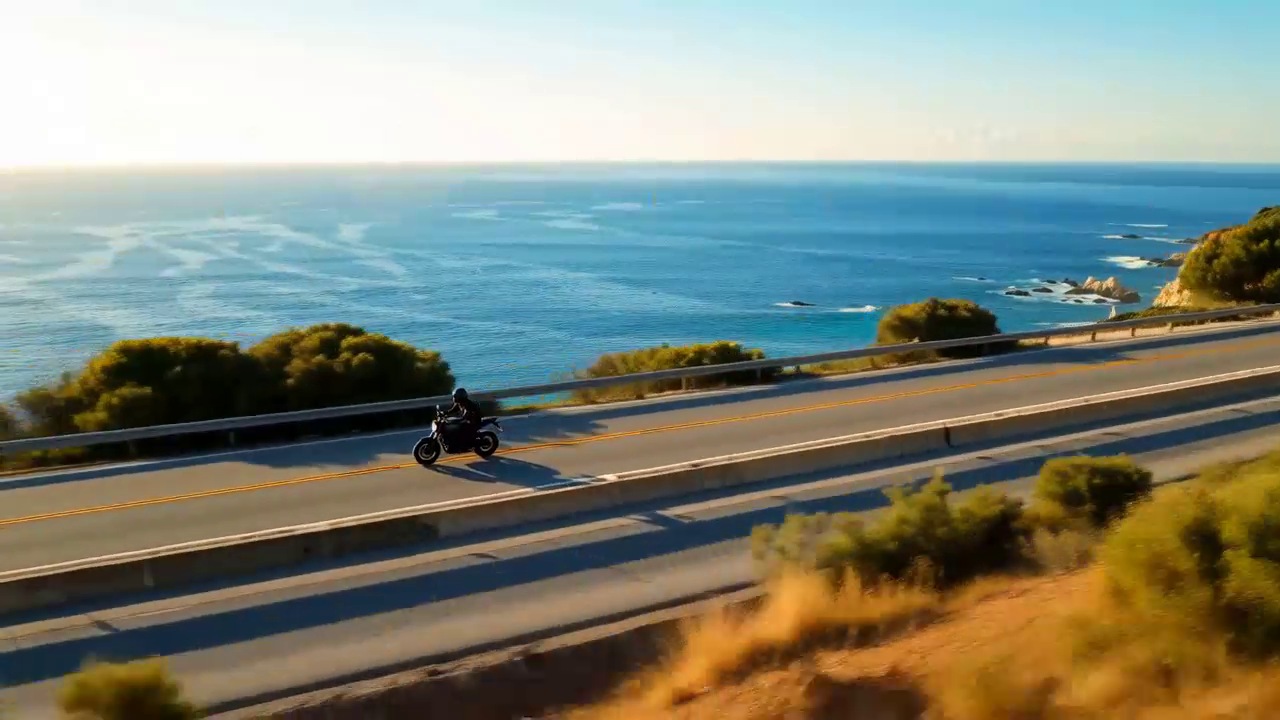}\gridimg{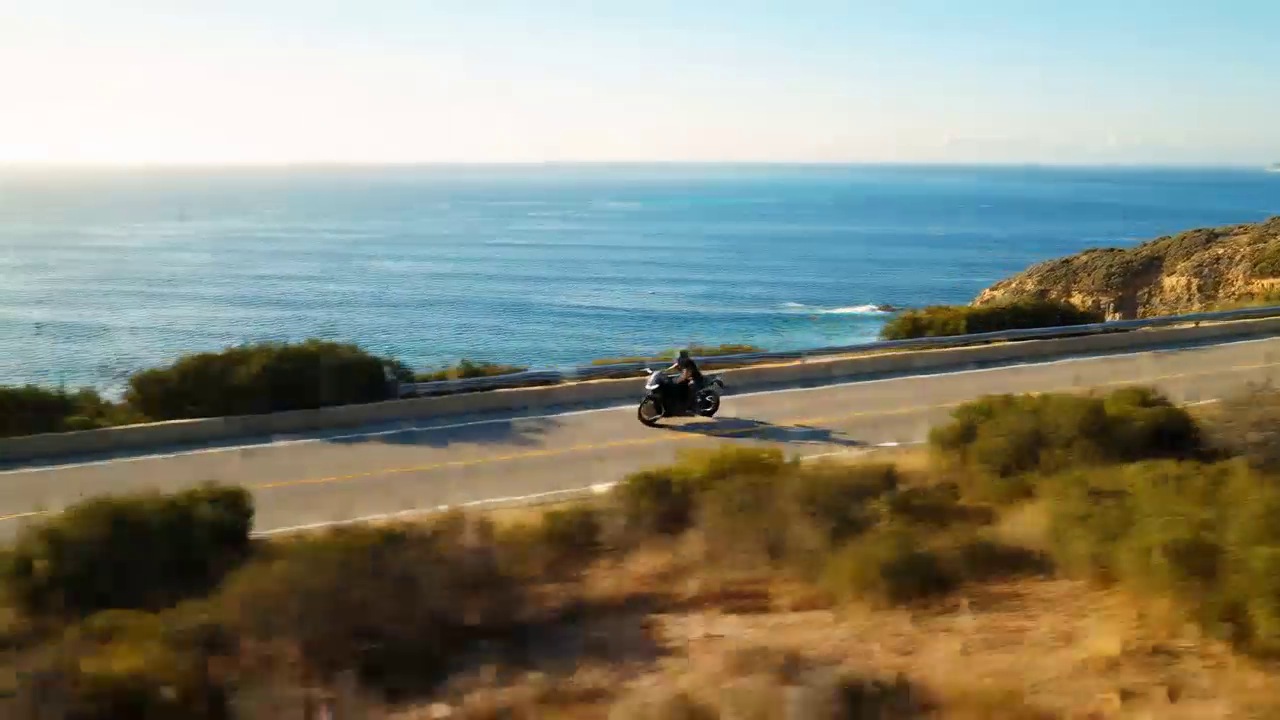}\gridimg{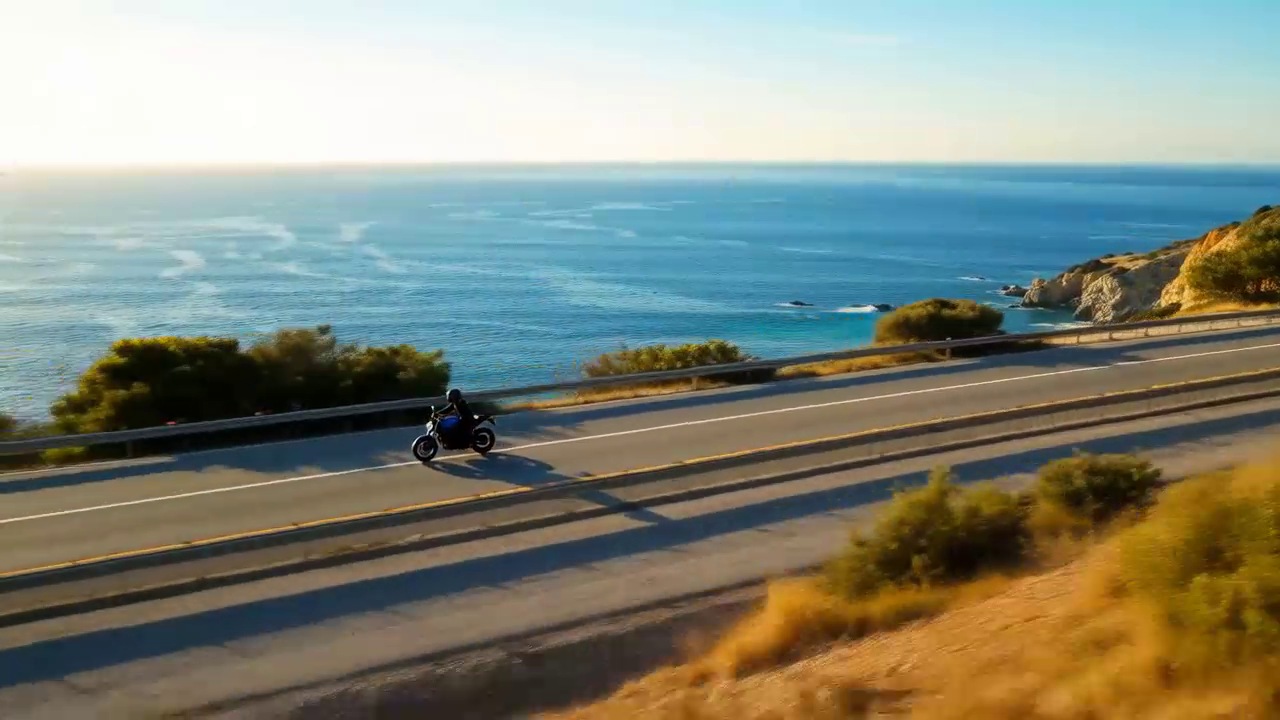}}
  }
  \caption{Wan2.2-T2V-A14B, frame $40$ of the cat (top) and motorcycle (bottom) prompts with each
  self-attention kernel.}
  \label{fig:wan22frames}
\end{figure}

\begin{figure}[t]
  \centering
  \setlength{\gridw}{\dimexpr(\columnwidth-\gridlab)/5\relax}
  {\offinterlineskip
  \hbox{\makebox[\gridlab]{}\gridhead{BF16}\gridhead{MXFP8}\gridhead{MXFP6}\gridhead{MXFP4}\gridhead{hybrid (MXFP6 + MXFP4)}}\vspace{1pt}
  \hbox{\gridrow{HunyuanVideo}\gridimg{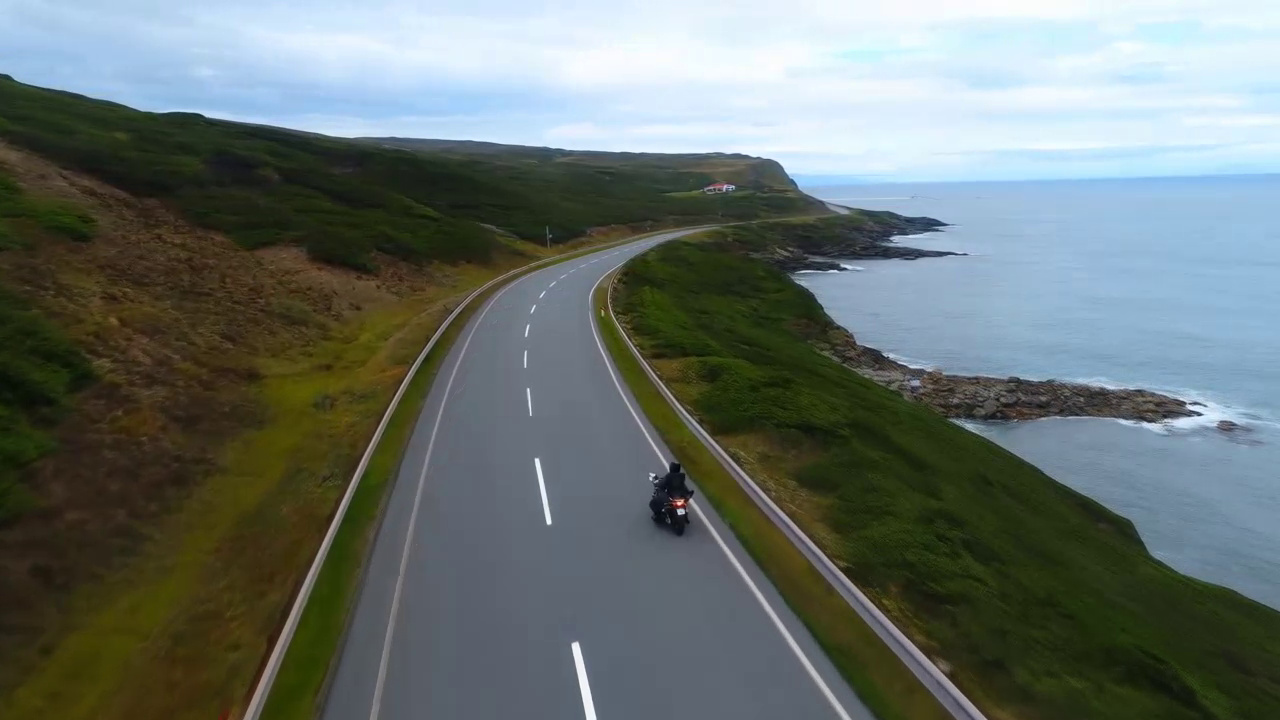}\gridimg{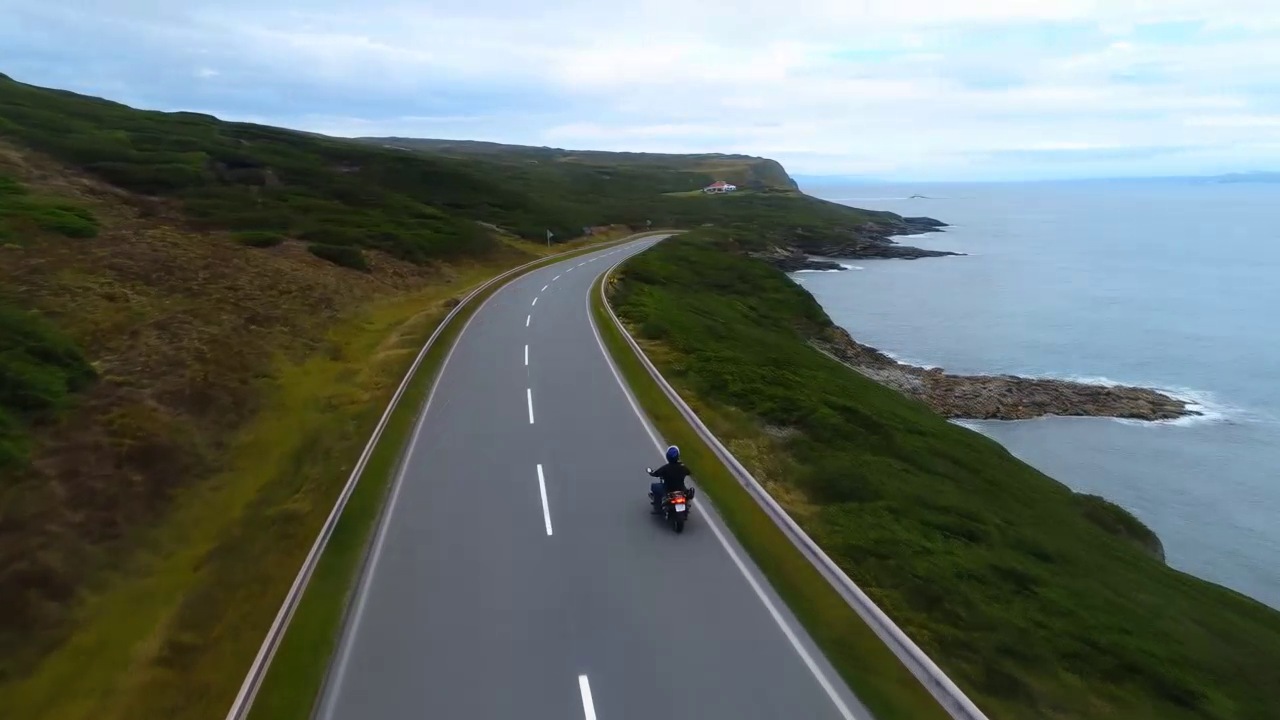}\gridimg{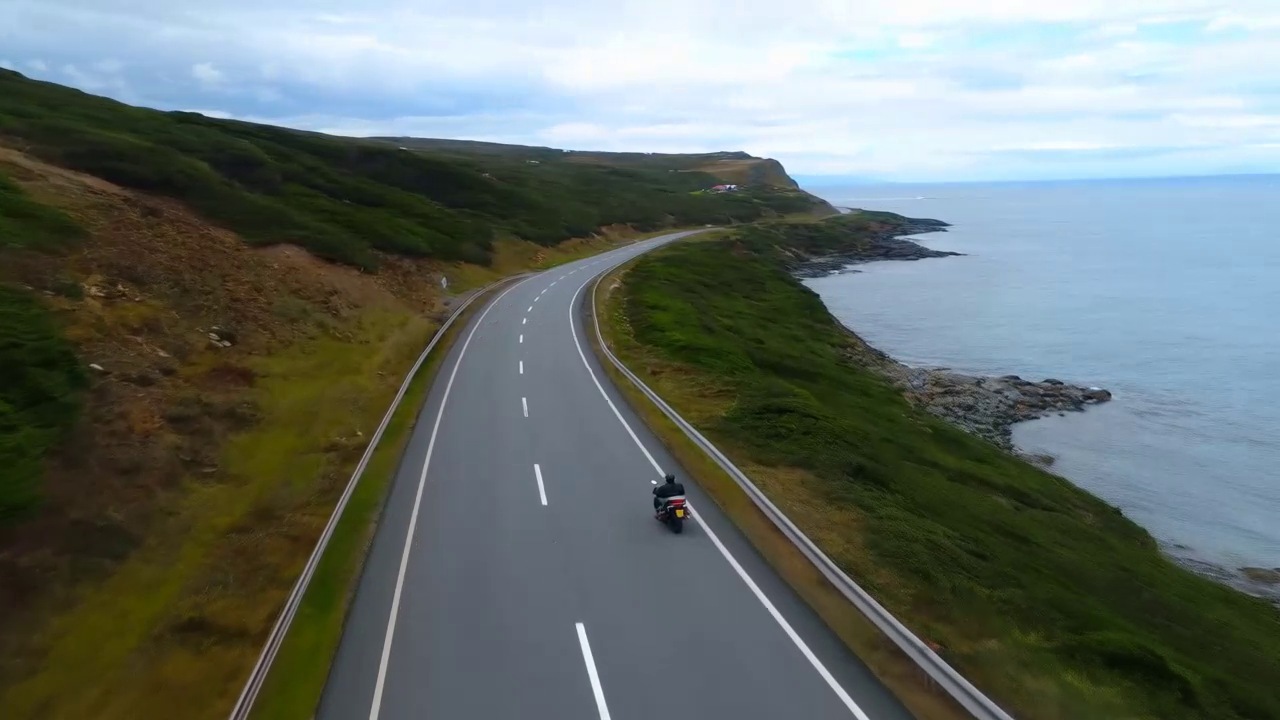}\gridimg{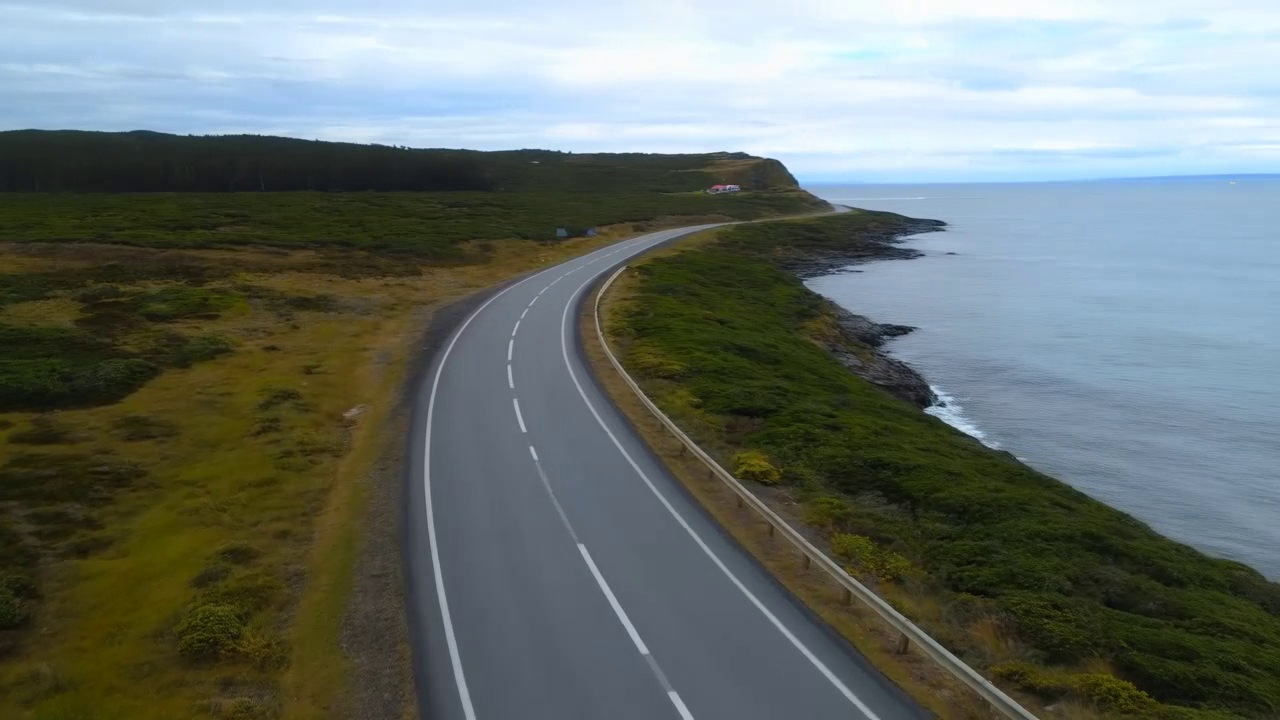}\gridimg{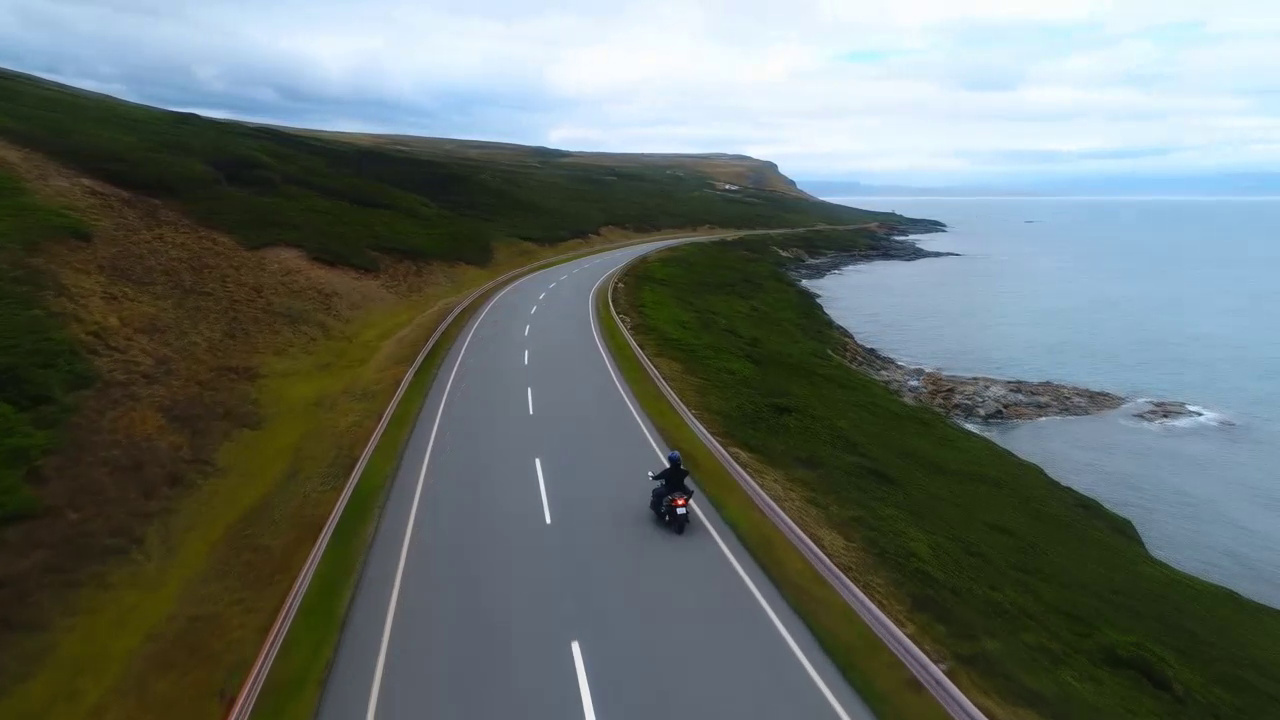}}
  \hbox{\gridrow{MiniMax-H3}\gridimg{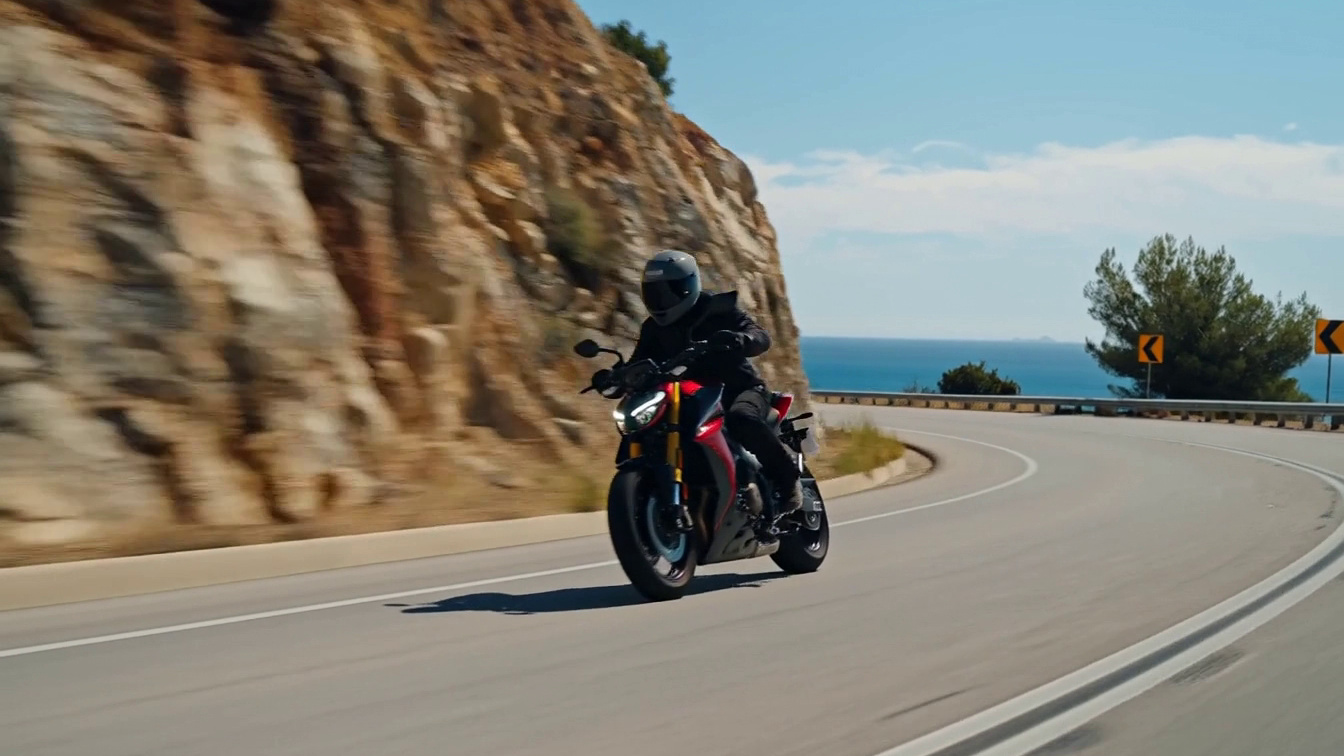}\gridimg{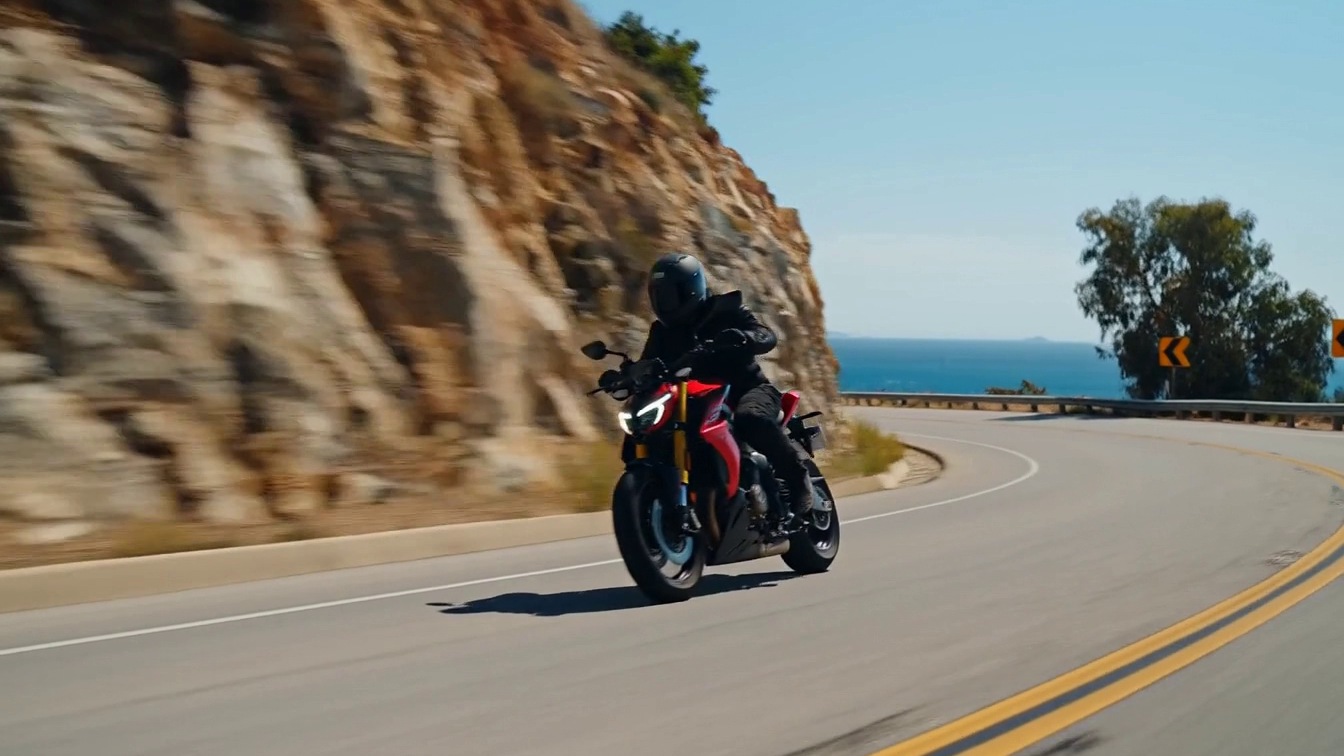}\gridimg{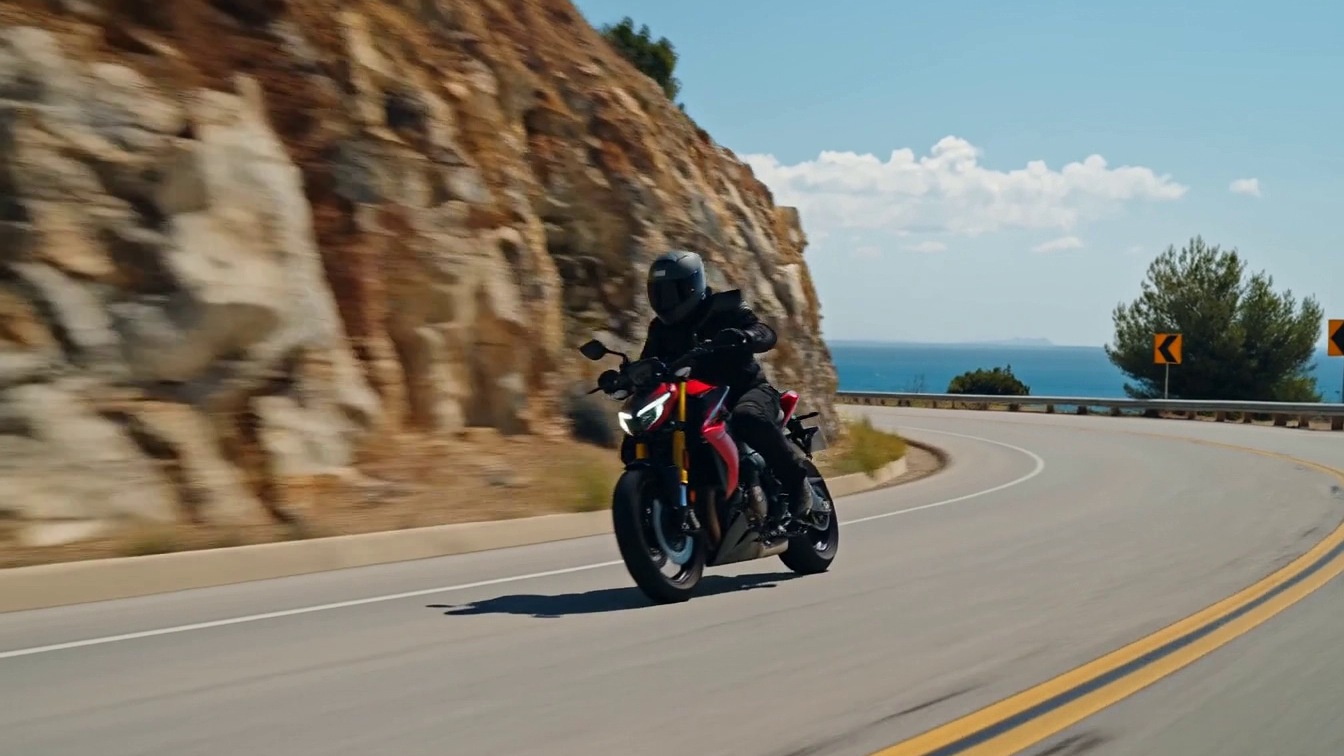}\gridimg{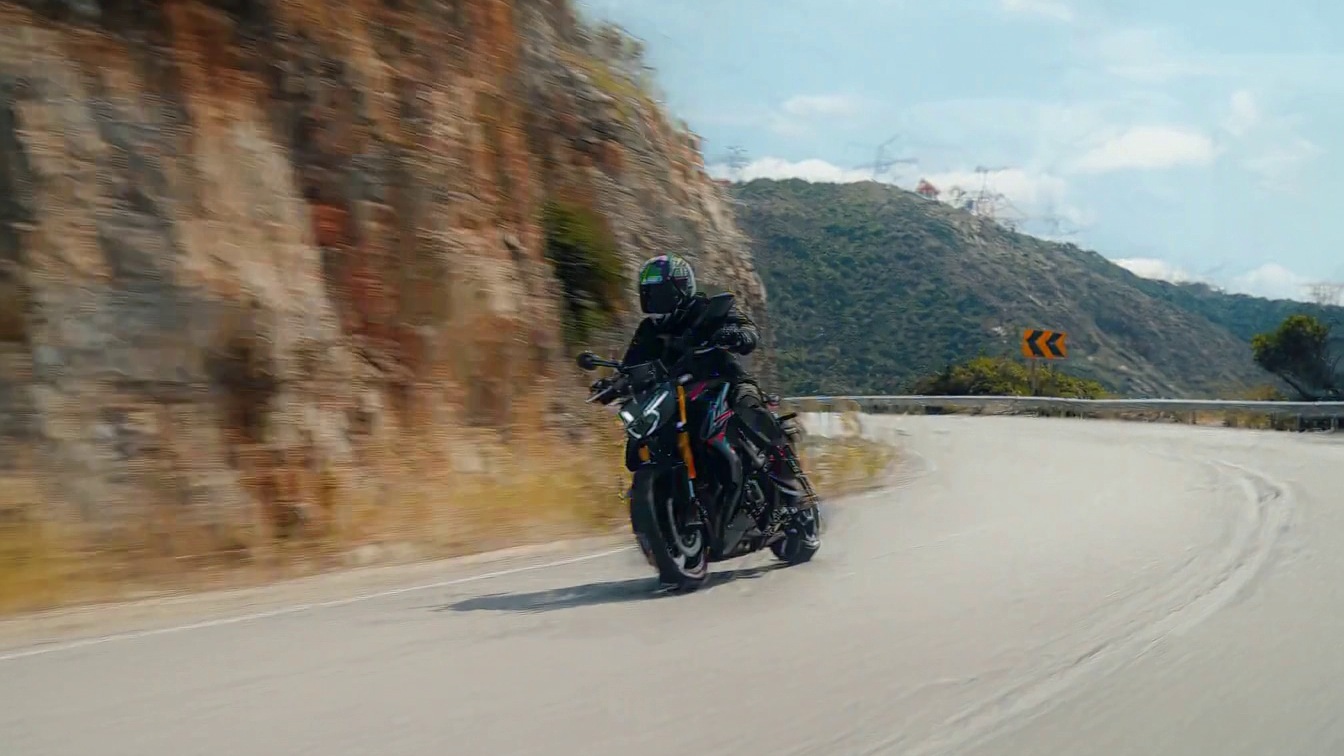}\gridimg{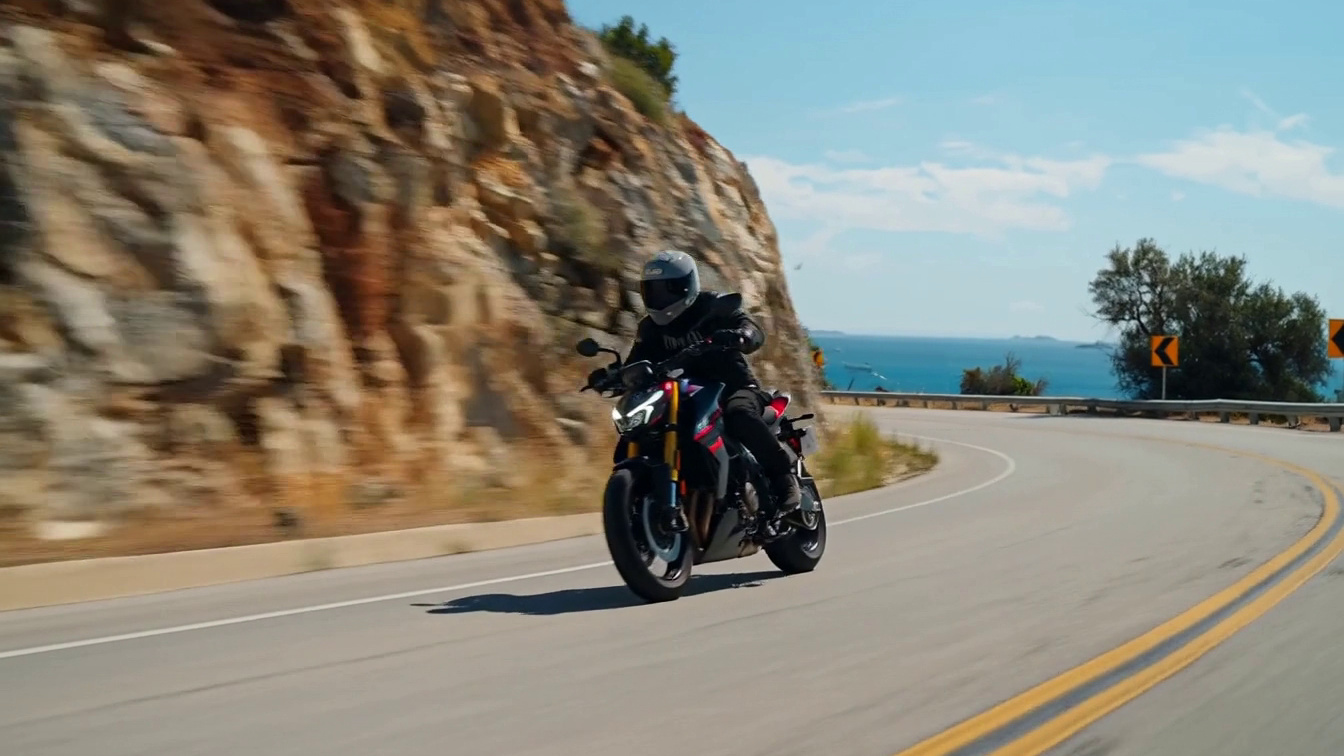}}
  }
  \caption{Motorcycle prompt across two additional models and all self-attention kernels.}
  \label{fig:othermodels}
  \vskip -0.1in
\end{figure}

\subsection{Other diffusion models}
\label{sec:othermodels}
We also replace self-attention with our kernels in HunyuanVideo~\cite{kong2024hunyuanvideo} and
MiniMax-H3~\cite{minimaxai2026h3} (Figure~\ref{fig:othermodels}).
The hybrid uses MXFP6 for the first and last steps (5 and 15 of 50 for
HunyuanVideo and MiniMax-H3) and MXFP4 otherwise. As in Wan2.2, MXFP4 changes the scene 
(the rider disappears in HunyuanVideo and the background
changes in MiniMax-H3); the hybrid, however, recovers the reference appearance.
MXFP8 and MXFP6 stay close to the reference, as in Wan2.2,
suggesting good generalization of our kernels across models.

\section{Conclusion}
\label{sec:conclusion}

We presented Fused-Softmax Attention, a quantized attention pipeline that addresses the growing
imbalance between matrix-core and vector-unit throughput. A speculatively frozen row maximum and an
affine exponential approximation make max subtraction and exponential evaluation foldable into the
score GEMM. A two-stage conversion then produces microscaled probabilities, reusing its block
reduction to derive their scales and detect overflow. Detected overshoots trigger an accurate rollback,
while evaluating the row sum as a matrix multiplication---sparse for MXFP8---removes the remaining
reduction from the vector path.

Implemented for the AMD CDNA\texttrademark{}~4 architecture, the resulting MXFP8, MXFP6 and MXFP4
kernels reach approximately 3.0, 4.0 and 4.6~PFLOP/s on an AMD Instinct\texttrademark{} MI355X GPU,
yielding speedups of $2.1\times$, $2.8\times$ and $3.2\times$ over BF16 attention. The additional error
from Fused-Softmax remains below the error floor set by quantizing the GEMM operands. In the cross-platform comparison with published related methods, our reported
throughput is $1.26$--$1.56\times$ higher at similar or higher cosine similarity.

These kernel gains translate to end-to-end diffusion inference. A hybrid MXFP6/4 schedule delivers
up to $2.1\times$ speedup on video generation while preserving the visual quality of the outputs.
Evaluations on Wan2.2-T2V-A14B, HunyuanVideo and MiniMax-H3 show a consistent
precision-quality pattern: MXFP8 and MXFP6 stay close to BF16, pure MXFP4 maximizes throughput at a
visible quality cost, and the hybrid schedule recovers that loss. 

The kernels are deployed: they ship in the open-source AITER library behind the \texttt{mha\_v4}
entry point~\cite{amd2025aiter}, where they compose with block-sparse attention. An earlier version of the hybrid kernel
powered AMD's MLPerf\textsuperscript{\textregistered} Inference v6.1 Wan2.2-T2V-A14B
submission~\cite{amd2026mlperf61blog,amd2026mlperf61rocm}, whose SingleStream latency fell
$1.7\times$ from the previous round on the same MI355X system.
Overall, our work shows that recasting softmax in terms of matrix operations enables
attention to capitalize on growing low-precision matrix throughput.

\bibliographystyle{abbrvnat}
\bibliography{fused_softmax_attention}

\clearpage
\appendix
\noindent{\Large\bf\scshape Appendix}
\vskip 0.1in
\suppressfloats[t] %

\section{VBench evaluation}
\label{app:vbench}
We follow the benchmarking protocol of the MLPerf\textsuperscript{\textregistered} Inference v6.1
text-to-video benchmark for Wan-2.2-T2V-A14B~\cite{reddi2020mlperf,mlcommons2026inference}. It
generates 720p videos ($81$ frames) in $20$ denoising steps with the UniPC scheduler~\cite{zhao2023unipc} and scores
them with VBench. Queries are issued by the open-source MLPerf\textsuperscript{\textregistered} LoadGen
library, in the Offline scenario for the VBench runs and the SingleStream scenario for the latency
measurement (Section~\ref{sec:e2e}).

We generate the 248 prompts of the VBench set~\cite{huang2024vbench} with Wan2.2-T2V-A14B ($81$ frames
at $1280\times720$, $20$ sampling steps, the reference guidance of $4$ for the high-noise expert and $3$ for the low-noise expert, the
reference negative prompt, one fixed seed and initial latent), once with BF16 attention and once with
each of our quantized kernels and the hybrid schedule, which runs MXFP6 attention for the first and last
three of the $20$ denoising steps and MXFP4 for the fourteen between them. The self-attention calls have
$s{=}75600$, $h{=}40$ and $d{=}128$. We run the BF16 Diffusers~\cite{vonplaten2022diffusers} checkpoint of the model through
xDiT~\cite{fang2024xdit} with \texttt{torch.compile}~\cite{ansel2024pytorch2}. GEMMs stay in BF16 and
cross-attention uses the BF16 attention kernel of AITER~\cite{amd2025aiter}, the
same kernel as the BF16 self-attention baseline, so only the self-attention kernel changes between the
runs.

\begin{figure}[t]
  \centering
  \includegraphics[width=\textwidth]{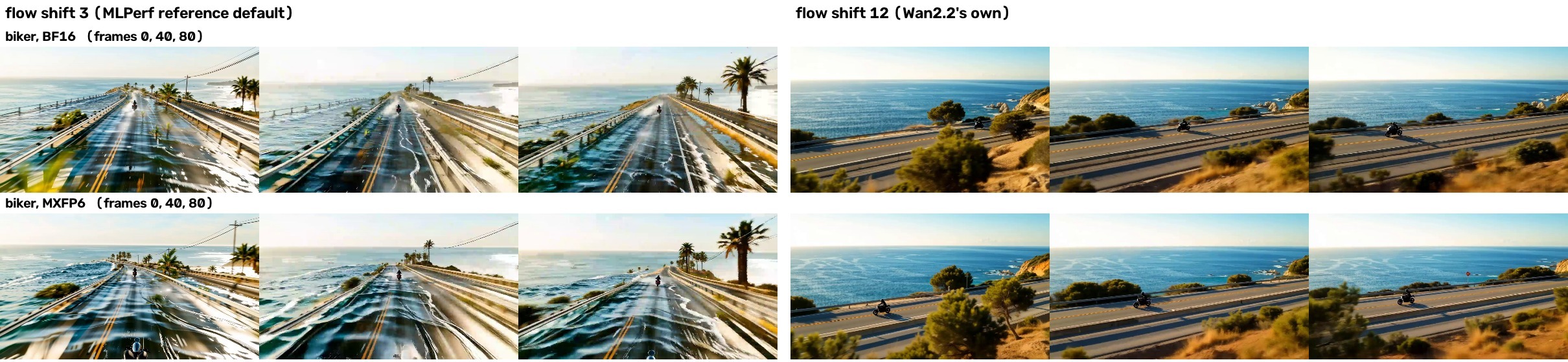}
  \caption{Wan2.2-T2V-A14B, motorcycle prompt, at the MLPerf text-to-video settings with the scheduler's default flow shift
  of $3$ (left) and with $12$ (right), frames $0$, $40$ and $80$, for BF16 and MXFP6 attention. Only the
  flow shift differs between the two sides.}
  \label{fig:flowshift}
\end{figure}

\paragraph{Flow shift.} The MLPerf reference implementation does not set the scheduler's flow shift, so
it inherits the checkpoint's default of $3$; Wan2.2's own configuration samples this model with a shift
of $12$~\cite{wan2025wan22repo}. At the default, the videos degrade even with BF16 attention:
Figure~\ref{fig:flowshift} shows the motorcycle prompt of Section~\ref{sec:visual} at both values, everything
else unchanged: the road melts into water at $3$ and stays intact at $12$. We
therefore use a shift of $12$ throughout, the only change from the reference settings.

\paragraph{Rollback capture.} For the per-call analysis of Section~\ref{sec:rollbackrates} we capture
the BF16 generations of the two prompts of the main paper's Figure~\ref{fig:wan22frames}, ``a motorcycle cruising along
a coastal highway'' and ``a cat grooming itself meticulously with its tongue''. We record the latent
that enters the transformer at each denoising step and replay those latents through the model with a
hook on its self-attention, so the captured calls are exactly those of the generation shown. Each
prompt gives $1600$ calls: all $20$ denoising steps, both classifier-free guidance passes and all $40$
layers.

\paragraph{Rollback on the worst calls.} The five calls with the highest rollback rates roll back on
$0.666$--$0.672\%$ of workgroup tiles. On them, enabling the gate changes the throughput by $-0.01\%$,
$-0.15\%$ and $+0.21\%$ for MXFP8, MXFP6 and MXFP4, respectively, all within run-to-run variation. Against the
model's BF16 attention output, our kernels reach a global relative error of $2.73\%$, $2.86\%$ and $18.03\%$,
against $2.74\%$, $2.95\%$ and $18.07\%$ for the conventional softmax on the same quantized operands (emulated),
averaged over the five calls. Over all $3200$ calls the median call rolls back on $0.014\%$ of workgroup tiles.

VBench scores each video automatically on six dimensions, reported here on a $0$--$100$ scale: subject and
background consistency (whether the subject and the background keep their appearance across frames),
motion smoothness, dynamic degree (how much motion the video contains), and appearance style and
scene (how well the video matches the style and scene named in the prompt). Some dimensions apply only
to a subset of the prompts. Table~\ref{tab:vbench} reports the mean of each dimension and the overall
score, the mean of the six.

\begin{table}[t]
\caption{VBench scores and closeness to the BF16 videos for Wan2.2-T2V-A14B, flow shift $12$. Each
VBench dimension is the mean over the prompts it applies to, on the $0$--$100$ scale; Overall is the mean
of the six dimensions, the benchmark's accuracy score. PSNR, SSIM and LPIPS compare every fourth
frame of each video with the same frame of the BF16 video of the same prompt, averaged over the 248
prompts.}
\label{tab:vbench}
\centering
\small\setlength{\tabcolsep}{4pt}%
\resizebox{\linewidth}{!}{%
\begin{tabular}{lrrrrrrr|rrr}
\toprule
Wan2.2 & \shortstack{Subject\\consistency} & \shortstack{Background\\consistency} & \shortstack{Motion\\smoothness} & \shortstack{Dynamic\\degree} & \shortstack{Appearance\\style} & Scene & Overall & PSNR & SSIM & LPIPS \\
\midrule
BF16   & 92.64 & 93.64 & 96.77 & 93.06 & 20.74 & 34.16 & 71.83 & \multicolumn{3}{c}{reference} \\
MXFP8  & 93.22 & 93.71 & 96.82 & 93.06 & 20.77 & 33.14 & 71.78 & 17.37 & 0.584 & 0.324 \\
MXFP6  & 93.03 & 93.88 & 96.72 & 91.67 & 20.58 & 35.25 & 71.85 & 16.32 & 0.546 & 0.363 \\
MXFP4  & 92.66 & 93.82 & 97.38 & 90.28 & 20.68 & 31.98 & 71.13 & 14.55 & 0.477 & 0.486 \\
hybrid & 93.09 & 94.01 & 97.06 & 93.06 & 20.42 & 34.45 & 72.01 & 16.31 & 0.540 & 0.382 \\
\bottomrule
\end{tabular}}
\end{table}

\begin{table}[t]
\caption{What each proxy width buys. $\mathrm{M}$ and $\mathrm{E}$ are the mantissa and exponent
widths of the pattern the proxy carries. The exponent-scale reach of $P$ is approximately
$2^{2^{\mathrm{E}-1}}$ and depends on $\mathrm{E}$ alone. The error
columns depend on $\mathrm{M}$ alone and are swept over one doubling interval of $f$; the last column is the
part of the RMS contributed by $\hat f$, separated in quadrature from the $1.770\%$ the ramp
carries regardless.}
\label{tab:proxywidth}
\centering
\normalsize\setlength{\tabcolsep}{6pt}%
\resizebox{\linewidth}{!}{%
\begin{tabular}{lrrrrrrr}
\toprule
 & & & reach & \multicolumn{3}{c}{relative error over one doubling interval of $f$} & added \\
\cmidrule(lr){5-7}
proxy & $\mathrm{M}$ & $\mathrm{E}$ & of $P$ & min & max & RMS & by $\hat f$ \\
\midrule
continuous          & $\infty$ & --- & ---       & $-3.939\%$ & $1.966\%$ & $1.770\%$ & --- \\
U32 / FP32  & 23 & 8 & $2^{128}$ & $-3.939\%$ & $1.966\%$ & $1.770\%$ & $0.000\%$ \\
U16 / FP16  & 10 & 5 & $2^{16}$  & $-3.972\%$ & $2.000\%$ & $1.770\%$ & $0.020\%$ \\
\textbf{U16 / BF16} & \textbf{7} & \textbf{8} & $\mathbf{2^{128}}$ & $\mathbf{-4.199\%}$ & $\mathbf{2.242\%}$ & $\mathbf{1.777\%}$ & $\mathbf{0.162\%}$ \\
U8 / E4M3  &  3 & 4 & $2^{8}$   & $-8.012\%$ & $6.398\%$ & $3.141\%$ & $2.595\%$ \\
\bottomrule
\end{tabular}}
\end{table}

\section{Proxy and Schraudolph accuracy}
\label{app:proxywidth}

Section~\ref{sec:2stage_cvt} narrows the Schraudolph proxy from the classic 32-bit FP32 pattern to
a 16-bit BF16 pattern. This appendix measures what the narrower mantissa costs in accuracy, and
Table~\ref{tab:proxywidth} sets that cost against the reach each candidate format affords. 

The Schraudolph approximation emulates the exponential operation by exponentiating the exponent bits while linearly interpolating the mantissa bits. For emulating $2^x$, there are two main error sources:
\begin{itemize}
  \item \emph{linear ramp}: the linear interpolation of mantissas versus exponentiating the mantissas.
  \item \emph{pattern quantization}: representing the interpolated result in the corresponding floating-point format by rounding.
\end{itemize}

\begin{figure}[h]
\centering
\begin{tikzpicture}
\begin{axis}[
  width=0.62\columnwidth, height=5.0cm, scale only axis,
  xlabel={$x$}, ylabel={$2^{x}$},
  domain=0:1, samples=200, xmin=0, xmax=1, ymin=0.95, ymax=2.05,
  xtick={0,0.25,0.5,0.75,1}, ytick={1,1.25,1.5,1.75,2},
  tick label style={font=\scriptsize}, label style={font=\small},
  legend style={font=\scriptsize, at={(0.02,0.98)}, anchor=north west,
                draw=none, fill=none, row sep=-2pt},
  every axis plot/.append style={very thick},
]
  \addplot[black] {2^x};
  \addplot[blue,dashed] {1+x};
  \addplot[red] {2^(floor(x-\schreps))*(1+(x-\schreps)-floor(x-\schreps))};
  \legend{true $2^{x}$, ramp\, ($\epsilon=0$), ramp\, ($\epsilon\approx0.058$)}
  \draw[<->,black] (axis cs:0.4427,1.3604) -- (axis cs:0.4427,1.4427);
  \node[anchor=north,font=\scriptsize,inner sep=1.5pt] at (axis cs:0.4427,1.345) {$6.1\%$};
\end{axis}
\end{tikzpicture}

\vspace{3pt}

\begin{tikzpicture}
\begin{axis}[
  width=0.62\columnwidth, height=4.2cm, scale only axis,
  xlabel={$x$}, ylabel={rel.\ error (\%)},
  domain=0:3, samples=600, xmin=0, xmax=3, ymin=-5.6, ymax=8.6,
  xtick={0,1,2,3}, ytick={-3.939,0,1.966,6.148},
  yticklabels={$-3.94$,$0$,$1.97$,$6.15$},
  tick label style={font=\scriptsize}, label style={font=\small},
  legend style={font=\scriptsize, at={(0.98,0.98)}, anchor=north east,
                draw=none, fill=none, legend columns=2, row sep=-2pt},
  every axis plot/.append style={very thick},
]
  \addplot[gray,dotted,thin,domain=0:3,samples=2,forget plot] {1.966};
  \addplot[gray,dotted,thin,domain=0:3,samples=2,forget plot] {-3.939};
  \addplot[blue,dashed] {100*((2^(floor(x))*(1+x-floor(x)))/(2^x)-1)};
  \addplot[red] {100*((2^(floor(x-\schreps))*(1+(x-\schreps)-floor(x-\schreps)))/(2^x)-1)};
  \legend{$\epsilon=0$, $\epsilon\approx0.058$}
\end{axis}
\end{tikzpicture}
\caption{The ramp error, independent of any datatype. \emph{Top:} over one doubling interval the mantissa
field interpolates $2^{x}$ linearly, the chord from $(0,1)$ to $(1,2)$. \emph{Bottom:} the resulting relative error is periodic with
period one doubling.}
\label{fig:schraudolph_ramp}
\end{figure}
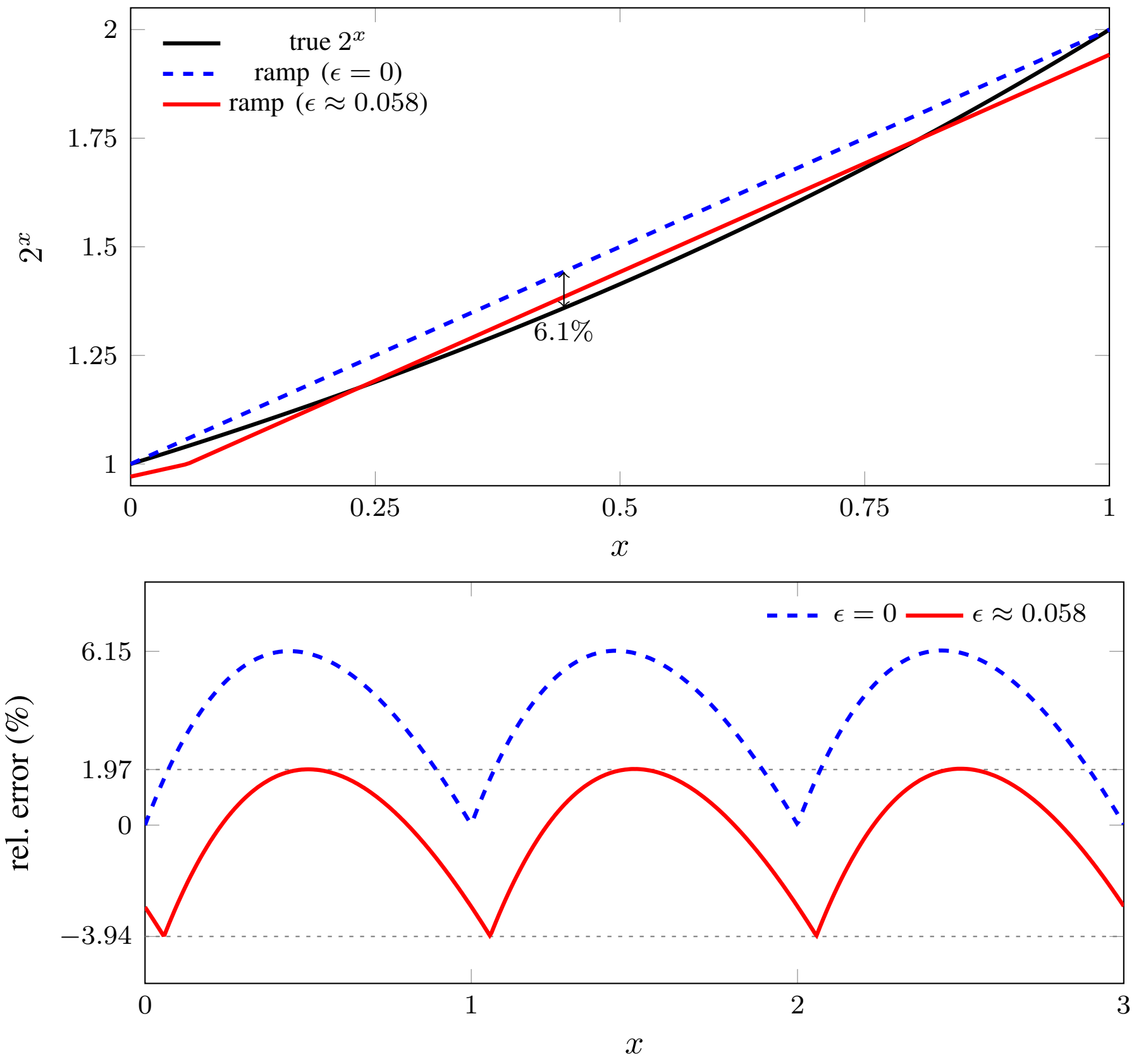

When it comes to the choice of proxy, the linear ramp is proxy-invariant: it is inherent to the Schraudolph approximation. Pattern quantization, by contrast, depends on the mantissa width of the datatype in question. Figure~\ref{fig:schraudolph_ramp} illustrates the linear ramp error over the range of input $x$.

Letting $f$ denote the fractional part of $x - \epsilon$, an $\mathrm{M}$-bit mantissa field cannot hold $f$ itself, only
$\hat f = \mathrm{round}(f\,2^{\mathrm{M}})/2^{\mathrm{M}}$, so the pattern carries $\hat f$ where
the argument still calls for $f$. That substitution is the entire second error source: $f$ is
discretized in steps of $2^{-\mathrm{M}}$, so the relative error it contributes is bounded by the
half-step $2^{-(\mathrm{M}+1)}$.

The half-step $2^{-(\mathrm{M}+1)}$ sets the scale. At $\mathrm{M}=10$ it is $0.049\%$, lost
beneath the ramp's $1.770\%$; at $\mathrm{M}=7$ it is $0.391\%$, adding $0.162\%$ in quadrature
for a $0.4\%$ relative rise in RMS; at $\mathrm{M}=3$ it reaches $6.250\%$ and dominates, lifting
RMS by $77\%$.

Even the $0.162\%$ has negligible measured output impact. $P$'s destination is FP6, whose own three
mantissa bits carry the same $6.250\%$ half-step, so the BF16 proxy is $16\times$ finer than the
format it feeds, and its rounding vanishes beneath stage~B's. Hence the design rule: the proxy
needs mantissa bits comfortably above the destination's, and $7$ against $3$ leaves four bits of
headroom, where an E4M3 proxy would carry \emph{exactly} three and quantize twice at the same
granularity.

Mantissa width is not the only constraint for the proxy. FP16 provides $10$ mantissa bits,
compared with $7$ for BF16 and $3$ for the U8/E4M3 proxy. However, FP16 and U8/E4M3 are
ruled out by their limited exponent range, since
$\mathrm{E}=5$ reaches only about $2^{16}$ and $\mathrm{E}=4$ about $2^{8}$, giving a far tighter
dynamic range than $\mathrm{E}=8$. Among the proxies that clear the range requirement, U32 is
marginally more accurate, but BF16 enables the packed two-stage conversion while adding only
$0.007$ percentage points to the RMS error. Therefore, the efficient choice is BF16.

\begin{table}[t]
  \caption{How \texttt{v\_pk\_maximum3\_f16} reads a \texttt{Stage-A} pattern. Here
  $e = \mathrm{pattern} \gg 7$ is the pattern's nine-bit sign-plus-exponent code, and
  $\Delta = e - 127$ its offset from the retained reference in doublings. The gate is an unsigned
  compare against the cap, $\Delta = 32$, of Section~\ref{sec:freeze}. Only the third band misreads;
  its upper boundary falls inside $e=504$, so the bands are delimited by pattern.}
  \label{tab:f16bands}
  \centering
  \small\setlength{\tabcolsep}{4pt}%
\resizebox{\linewidth}{!}{%
\begin{tabular}{llllll}
  \toprule
  $e$ & pattern & $\Delta$ & read as FP16 & reduction & gate \\
  \midrule
  $0$--$247$ & \texttt{0x0000}--\texttt{0x7BFF} & $<121$ & positive finite
    & correct & fires iff above the cap \\
  $248$--$255$ & \texttt{0x7C00}--\texttt{0x7FFF} & $121$--$128$ & $+\infty$ or $+$NaN
    & kept ($\infty$) or propagated (NaN) & fires \\
  $\mathbf{256}$--$\mathbf{504}$ & \textbf{\texttt{0x8000}--\texttt{0xFC00}}
    & $\mathbf{129}$--$\mathbf{377}$ & \textbf{negative finite, and $-\infty$}
    & \textbf{dropped} & \textbf{misses} \\
  $504$--$511$ & \texttt{0xFC01}--\texttt{0xFFFF} & $377$--$384$ & $-$NaN
    & propagates NaN & fires \\
  \bottomrule
  \end{tabular}}
\end{table}

\section{Evaluation on the $16$-bit maximum operations}
\label{app:f16max}

Section~\ref{sec:2stage_cvt} mentions placing the block maximum after \texttt{Stage-A}, on packed
data. Two $16$-bit reductions can do that. \texttt{v\_pk\_max\_u16} reads the codes as the
unsigned integers they are and is exact; \texttt{v\_pk\_maximum3\_f16} retires twice as many
values per issue but hands a BF16 pattern to an FP16 interpretation, and two properties of the
construction decide whether a gate built on it is sound. Our MXFP6 kernel takes the exact
reduction throughout; the MXFP4 one takes \texttt{maximum3} in the steady state and falls back to
the unsigned reduction in the rollback.

\paragraph{The unsigned reduction is exact over every doubling.}
The packed normalized conversion is monotone in its affine input, so unsigned code order preserves
input order across the whole code space: no peak can be dropped, and there is
no \texttt{Inf}/NaN band to invent one. The pattern domain also makes the gate and the rollback
integer work, one doubling being exactly $128$ pattern units. Its only ceiling belongs to the
conversion rather than the reduction, since a coordinate saturated at \texttt{0xFFFF} is
unrecoverable from its code. The price is issue slots: $15$ two-input maxima and a halfword fold
per $32$-element block, $17$ instructions against the $10$ of a \texttt{maximum3} chain.

\paragraph{\texttt{Stage-A} cannot emit anything but a pattern.}
The packed normalized conversion clamps to $[0,1]$ before scaling by $65535$, so the code written
out always lies in $[0,65535]$; a coordinate past the top saturates at \texttt{0xFFFF} rather than
wrapping to a small one. Below the sign bit, ordering survives the reinterpretation: FP16's
$1{+}5{+}10$ split ranks positive codes exactly as BF16's $1{+}8{+}7$ does. What FP16 adds is an
\texttt{Inf}/NaN band above \texttt{0x7C00}, and it costs nothing, since the instruction
propagates NaN and every FP16 NaN pattern is at least \texttt{0x7C00} unsigned, hence above any
useful cap. The one unsigned compare that catches a genuine overshoot therefore also catches a
poisoned reduction, after which the rollback re-reduces with \texttt{v\_pk\_max\_u16}.

\paragraph{A large overshoot reaches the sign bit.}
Schraudolph writes the score into the exponent field, so an overshoot climbs through the exponent
into bit~15 rather than saturating. Both formats read bit~15 as a sign but place their
\texttt{Inf}/NaN bands at different heights, which opens a gap. Of the four bands in
Table~\ref{tab:f16bands} only the third misreads: a pattern there is an ordinary negative finite
to FP16, so \texttt{maximum3} has nothing to propagate and returns the positive neighbor, and the
peak leaves the reduction before the gate sees it. $-\infty$ (\texttt{0xFC00}) belongs to that
band, since only NaN propagation is sign-agnostic.

What matters is that the band is \texttt{Inf}- and NaN-free when interpreted as BF16. \texttt{Stage-B} reads the dropped
pattern as BF16, where bit~15 is again a sign and the exponent sits far below the reference, so
the row's dominant probability is packed as a small negative weight: the mass is lost and the $PV$
MFMA and the $\ell$ sum take a finite perturbation of one row, with nothing to propagate. A NaN
reaching those accumulators would destroy the output irrecoverably, and that is the failure this
band cannot produce. A missed gate would instead cause a finite perturbation; we observed no such
missed-gate failure across the evaluated model calls. The band is also hard to reach, opening at $\Delta = 129$ against a cap at
$\Delta = 32$ and only when the peak is isolated within its own $32$-element block --- any
block-mate above the cap but below $129$ still fires the gate. \texttt{maximum3} therefore
guarantees $97$ doublings rather than the whole code space, which is why MXFP6 pays the seven extra
issues per block and only MXFP4 takes the cheaper reduction.

\FloatBarrier
\section{Memory bandwidth roofline}
\label{app:memroofline}

The operands have to reach the matrix cores before they can be multiplied, over two hops:
high-bandwidth memory (HBM) into LDS, and LDS into registers. We cost each against the matrix and vector work of Section~\ref{sec:motivation}.

\paragraph{HBM to LDS.}
Without cache reuse, each $K_j/V_j$ loads $2Nd$, $1.5Nd$, or $Nd$ bytes at
FP8, FP6, or FP4 through the HBM. Sharing $8$\,TB/s across $256$ CUs at $2.4$\,GHz gives
$13.02$\,B/clk/CU, hence
\[
 (T^{FP8}_{\mathrm{HBM}},T^{FP6}_{\mathrm{HBM}},T^{FP4}_{\mathrm{HBM}})
 \approx(2517,1887,1258)\ \mathrm{clk}.
\]
We double-buffer LDS: slot 0 supplies registers for the current tile while
slot 1 receives the next tile, then their roles swap. Both wave groups use
the same $K_j/V_j$, so the next transfer arrives in time when
\[
 T_{\mathrm{HBM}}<T_{\mathrm{GEMM}}+T_{\mathrm{VALU}}.
\]
Even without cache reuse, these transfer costs remain below the combined matrix and vector work,
and any reuse lowers them further.

\paragraph{LDS to registers.}
Each of the eight waves reads $K_j/V_j$ from LDS. At $256$\,B/clk/CU
\cite{amd2025cdna4whitepaper}, the ideal read costs are
\[
 \frac{8(2Nd,\,1.5Nd,\,Nd)}{256}
 =(1024,768,512)\ \mathrm{clk}
\]
for FP8, FP6, and FP4, respectively. These are also below the matrix cycle cost, and we use this to our advantage by interleaving the
LDS reading into the matrix execution shadow.

\end{document}